\pdfoutput=1

\documentclass[11pt]{article}

\usepackage[final]{acl}

\usepackage{times}
\usepackage{latexsym}

\usepackage[T1]{fontenc}

\usepackage[utf8]{inputenc}

\usepackage{microtype}

\usepackage{inconsolata}

\usepackage{graphicx}
\usepackage{booktabs}
\usepackage{multirow}
\usepackage{color}
\usepackage{amsmath}
\usepackage{enumitem}

\usepackage{pifont}
\usepackage{bbding}
\usepackage{fontawesome}

\title{A Silver Bullet or a Compromise for Full Attention? \\ A Comprehensive Study of Gist Token-based Context Compression}

\author{
    Chenlong Deng$^{1,2\dagger}$, Zhisong Zhang$^{2*}$, Kelong Mao$^1$, \\
    \textbf{Shuaiyi Li$^2$, Xinting Huang$^2$, Dong Yu$^2$, Zhicheng Dou$^{1*}$} \\
    $^1$Gaoling School of Artificial Intelligence, Renmin University of China \\ 
    $^2$Tencent AI Lab\\ 
    \texttt{\{dengchenlong,dou\}@ruc.edu.cn} \\
    \texttt{zhisonzhang@tencent.com}
}

\begin{document}
\maketitle
\renewcommand{\thefootnote}{\fnsymbol{footnote}} 
\footnotetext[2]{This work was done during internship at Tencent AI Lab.} 
\footnotetext[1]{Corresponding authors.}
\renewcommand{\thefootnote}{\arabic{footnote}} 

\begin{abstract}
In this work, we provide a thorough investigation of gist-based context compression methods to improve long-context processing in large language models. We focus on two key questions: (1) How well can these methods replace full attention models? and (2) What potential failure patterns arise due to compression? Through extensive experiments, we show that while gist-based compression can achieve near-lossless performance on tasks like retrieval-augmented generation and long-document QA, it faces challenges in tasks like synthetic recall. Furthermore, we identify three key failure patterns: lost by the boundary, lost if surprise, and lost along the way. To mitigate these issues, we propose two effective strategies: fine-grained autoencoding, which enhances the reconstruction of original token information, and segment-wise token importance estimation, which adjusts optimization based on token dependencies.
Our work provides valuable insights into the understanding of gist token-based context compression and offers practical strategies for improving compression capabilities.
\end{abstract}

\section{Introduction}
Large language models (LLMs) are increasingly recognized as a key pathway toward general artificial intelligence~\cite{GPT4-report, llm-survey}, with long-context processing emerging as a critical research frontier~\cite{PI, YaRN}. This capability is crucial for advanced applications like retrieval-augmented generation (RAG), long-term memory systems, and complex reasoning frameworks~\cite{rag-survey, irllm-survey, long-term-survey, CoT, verify-step-by-step}.
Despite the proliferation of architectural innovations, Transformer-based models remain the performance standard. However, these architectures face significant computational challenges when processing extended text sequences: the key-value (KV) cache memory grows linearly with sequence length, while the attention mechanism's quadratic computational scaling introduces substantial overhead. In models like Llama3-8B~\cite{llama3}, a 128K context KV cache can consume memory equivalent to the entire model's parameters, limiting deployment on edge devices and constraining context windows.

\begin{figure*}[!t]
	\centering
	\includegraphics[width=0.95\linewidth]{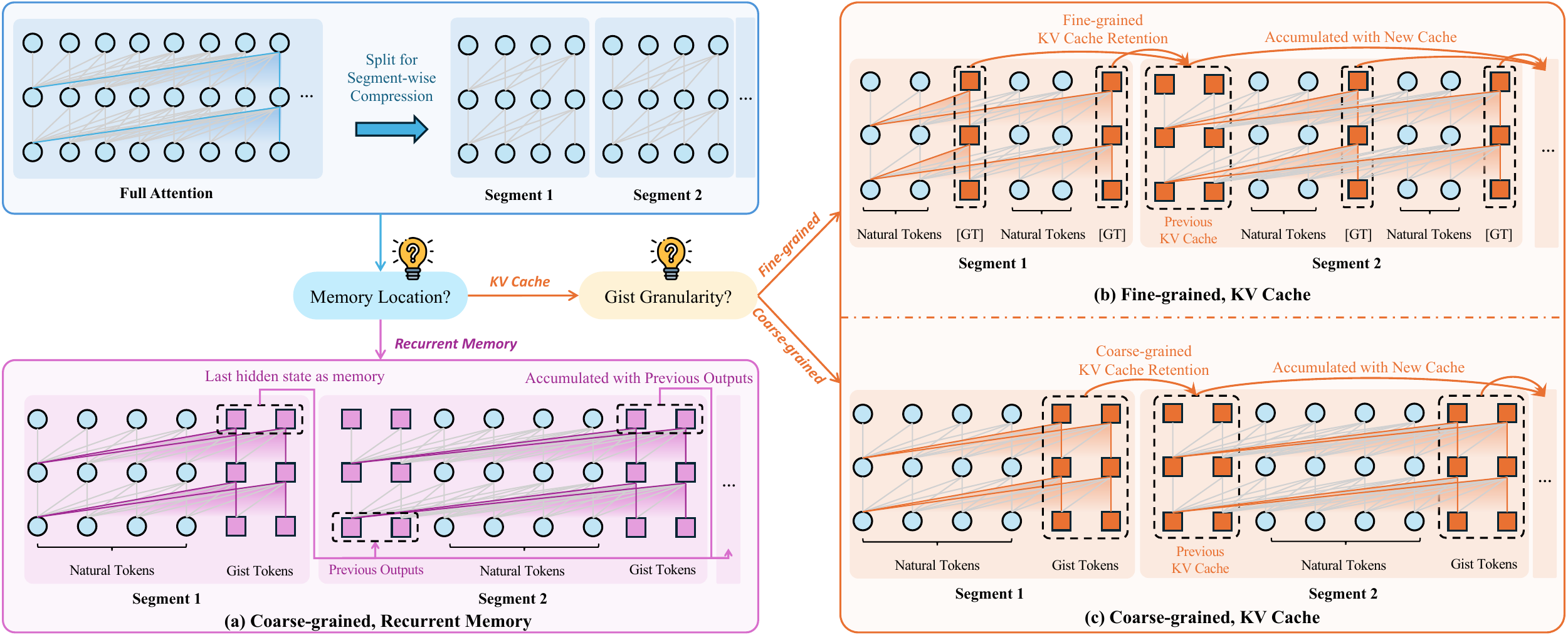}
	\caption{Overview of gist token-based context compression architectures. Long texts are segmented for compression, enabling diverse architectures through different \textit{memory locations} and \textit{gist granularity}.}
	\label{fig: taxonomy}
    \vspace{-2ex}
\end{figure*}

A promising approach to mitigate these challenges involves reducing overhead by compressing the number of past tokens stored in the KV cache.
This work focuses on a specific type of compression method that condenses the context into a small set of special tokens, called gist tokens~\cite{Gist}.\footnote{Previous works refer to this concept by various names. We unify these terms and refer to them as ``gist tokens'' for consistency in this paper.} By replacing the original tokens with a limited number of gist tokens, these methods effectively reduce both KV cache size and computational cost. While such techniques have been successfully applied in real-world tasks~\cite{memorag}, two critical questions remain unresolved:

\textit{Q1}: To what extent can this architecture replace full attention models? 
\textit{Q2}: Does the compression introduce potential, yet significant, failure patterns?

In this work, we thoroughly investigate these two questions through extensive experiments.  
Specifically, we propose a unified framework for categorizing existing gist-based model architectures along two dimensions: \textit{Memory Location} and \textit{Gist Granularity}. We provide comprehensive evaluations for them with a wide range of language tasks.

For \textit{Q1}, our findings indicate that the fine-grained KV cache architecture (referred to as \textit{Fine KV}) is highly effective, achieving near-lossless compression performance on various tasks, such as RAG, long-document QA, and summarization, when compared to the full attention model. However, it still exhibits notable gaps in tasks like reranking and synthetic recall, suggesting that while promising, it is prone to severe compression failures in certain scenarios. Regarding \textit{Q2}, we conduct a probing experiment focused on context reconstruction and discover that the compression bottlenecks occur in the gist representations. We further identify three failure patterns resulting from this bottleneck: 1) \textit{lost by the boundary}, where generation degrades near the start of a segment; 2) \textit{lost if surprise}, where unexpected details tend to be ignored if budgets are limited; and 3) \textit{lost along the way}, where compressed models make errors midway for tasks requiring precise recall.

Building on the above findings, we further propose two strategies to enhance the Fine KV architecture for more effective context compression.
The first, \textit{fine-grained autoencoding}, adds a weak decoder with an autoencoding loss to reconstruct original token information from gist tokens, ensuring efficient and accurate compression. 
The second, \textit{segment-wise token importance estimation}, adjusts loss weights based on a token’s dependency on the compressed context, dynamically optimizing tokens that require more contextual understanding.
Experiments show that both strategies significantly improve model performance, with joint optimization achieving the best results.

The contributions of this work are:
\begin{itemize}[leftmargin=*]
\vspace*{-2mm}
\item We propose a unified framework for categorizing existing gist-based model architectures and conduct comprehensive experiments to evaluate their effectiveness. (\S\ref{sec:category}) 
\vspace*{-2mm}
\item We show that that gist-based models achieve near-lossless performance on many tasks but still face challenges in particular scenarios. (\S\ref{sec: overall evaluation})
\vspace*{-2mm}
\item We identify three critical failure patterns arising from compression bottlenecks, offering valuable insights into the limitations of current gist-based compression methods. (\S\ref{sec: understand why and how})
\vspace*{-2mm}
\item We propose two strategies: fine-grained autoencoding and segment-wise token importance estimation, which effectively mitigate these bottlenecks and enhance model performance. (\S\ref{sec:mitigating})
\end{itemize}

\section{Preliminaries}
\label{sec:category}

Gist token-based context compression reduces KV cache by using some special tokens, which are referred to as gists, to represent the full context. The number of special tokens is much fewer than that of the full context, leading to lower memory usage. While many pervious work studies compressing the full prompt at once~\cite{Gist, ICAE}, we focus on a generalized scenario that dynamically compresses and generates context on the fly, as such setting holds promise for broader general-purpose tasks. To this end, we provide a unified perspective to analyze and understand existing architectures.

Figure~\ref{fig: taxonomy} illustrates an overview of gist-based context compression methods. We take a segment-wise approach that splits the input sequence into segments and iteratively applies compression for each segment. Assuming an input sequence $X=\lbrack x_1, \ldots, x_n\rbrack$, it is divided into segments of fixed length $L$, where the $i$-th segment is represented as $S_i = \lbrack x_{(i-1)\cdot L+1}, \ldots, x_{(i-1)\cdot L+L}\rbrack$. When processing the $i$-th segment, the model accumulates all previously compressed information and generates new compressed representations as the memory for later processing:
\begin{align*}
    \hat{G}_{< (i+1)} \leftarrow \text{LLM}([\hat{G}_{< i}, \text{Insert}(S_i, G_i)])
\end{align*}
Here, $G_i = \lbrack g_1, \ldots, g_t\rbrack$ are new gist tokens inserted into the $i$-th segment, and $\hat{G}_i$ are compressed context representations preceding this segment. The function $\text{Insert}(\cdot)$ denotes the insertion of gist tokens into the input sequence. This procedure effectively compresses the information of $L$ tokens into $t$ tokens, achieving a compression ratio of $L/t$. For example, with a compression ratio of 4, every four raw tokens can be replaced by one gist token on average, thereby reaching a 75\% reduction in memory usage. Following this formula, existing architectures can be categorized along two dimensions: ``memory location'' and ``gist granularity''.

\paragraph{Memory Location} 
After the forward pass of each segment, we can choose to store either the last hidden states of the gist tokens or their KV cache as memory. Opting for the last hidden states is commonly referred to as ``recurrent memory'', which serves as input embeddings to deliver compressed context to subsequent segments. Note that this design can be viewed as a segment-wise RNN, and typical representatives include RMT~\cite{RMT} and AutoCompressors~\cite{AutoCompressors}. Alternatively, the KV cache of the gist tokens can be directly reused as the memory to avoid extra computations, and this shares the same design as in sparse attention. Typical representatives of the KV approach include Gist~\cite{Gist}, Landmark~\cite{Landmark} and Activation Beacon~\cite{Beacon}.

\paragraph{Gist Granularity} 
The $\text{Insert}(\cdot)$ function in the formula can be implemented in two ways: (1) Coarse-grained: Gist tokens are appended after all raw tokens, allowing each gist token to attend to the entire segment and all preceding contexts, which is the scheme adopted in most previous works; (2) Fine-grained: Gist tokens are evenly inserted among the raw tokens, enabling each gist token to focus on a specific context, which is investigated in Activation Beacon~\cite{Beacon}. Besides, this design can also enhance language modeling through an implicit chain-of-thought mechanism. 

Notably, the combination of recurrent memory and fine-grained gist tokens is practically infeasible, since it requires too many non-parallelizable forward passes within a segment. Therefore, we mainly explore the remaining three combinations in this work, as illustrated in Figure~\ref{fig: taxonomy}.

\section{Can Gist Tokens Replace Full Attention in an Efficient and Effective Way?}
\label{sec: overall evaluation}

\begin{figure*}[!t]
	\centering
	\includegraphics[width=\linewidth]{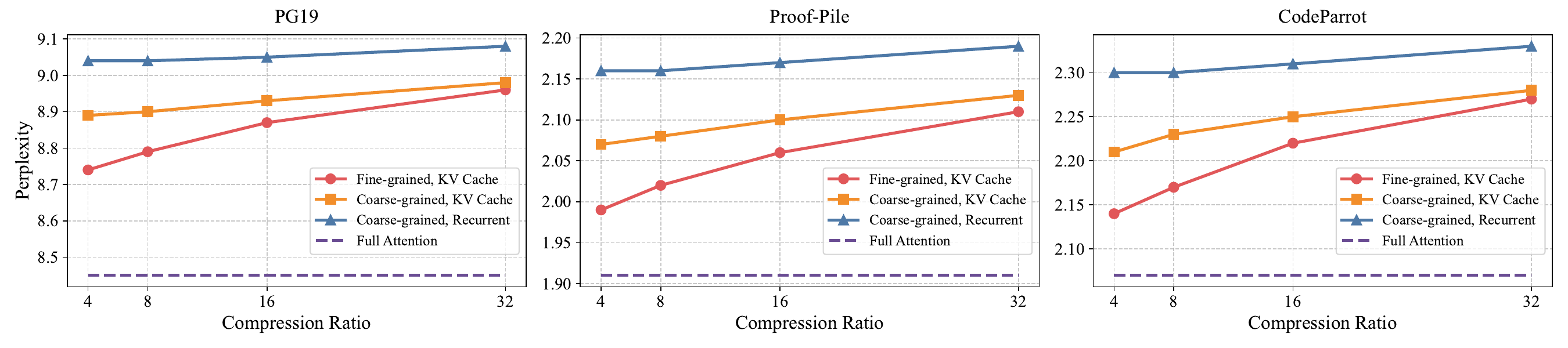}
	\caption{Comparisons of different compression methods on perplexity evaluation for language modeling.}
	\label{fig: Overall-PPL}
\end{figure*}

\subsection{Experimental Setup} 
\paragraph{Training Recipes} 
In our main experiments, we perform continued-training on the base models using a general-purpose corpus to analyze their intrinsic context compression capabilities. To avoid potential confounding effects from techniques like supervised fine-tuning, we focus exclusively on the base models rather than the SFT ones.\footnote{Extra analysis of SFT is showed in Appendix~\ref{appendix: sft results}.} Specifically, we select Llama3.1-8B~\cite{llama3} and Qwen2-7B~\cite{Qwen2} as our base models, given their widespread recognition and adoption in the community. We use the SlimPajama dataset and follow the processing procedure of \citet{LongContextDataEngineering}, by upsampling long sequences and ultimately obtaining 3B tokens for training. 
Further training details are provided in Appendix~\ref{appendix: training details}.

\paragraph{Evaluation Tasks} We perform extensive experiments, covering a wide range of tasks: \textbf{(1) Language modeling}, for which we evaluate perplexity on PG19~\cite{PG19}, Proof-Pile~\cite{Proof-Pile}, and CodeParrot~\cite{CodeParrot}; \textbf{(2) Weak Context-dependent Tasks},\footnote{These tasks do not inherently require long contexts. We increase their context length by adding demonstration examples, although the tasks themselves exhibit only weak dependence on this additional context.} for which we evaluate four tasks with MMLU-Pro~\cite{MMLU-Pro}, GSM8K~\cite{GSM8K}, HellaSwag~\cite{HellaSwag}, and BBH~\cite{BBH}, to evaluate the model's abilities in knowledge, mathematics, common sense, and comprehensive reasoning, respectively; \textbf{(3) Long Context Tasks}, which thoroughly assess the model's handling of long texts and we select seven types of tasks: RAG, Rerank, LongQA, Many-shot ICL, Synthetic Recall, Summarization, and Code. The datasets selected for testing these tasks include portions from popular long-text benchmarks such as RULER~\cite{RULER} and $\infty$Bench~\cite{Infbench}. Inspired by \citet{Helmet}'s setting, we adopt 2-shot demonstrations to ensure a robust evaluation of long-context performance. Further details on the datasets and metrics are provided in Appendix~\ref{appendix: evaluation details}.

\subsection{Overall Performance Comparisons}
We present the results of the Llama model in the main text, while the results of the Qwen model are presented in Appendix~\ref{appendix: qwen performance}.

\paragraph{Language Modeling} As shown in Figure~\ref{fig: Overall-PPL}, the differences between the architectures are clear and consistent across all datasets. Full attention outperforms all methods that compress contexts. Among the compression-enhanced architectures, fine-grained compression delivers better performance than coarse-grained, and KV cache performs better than recurrent memory. Note that the absolute differences in perplexity are small; for example, with a compression ratio of 4, the gap between the fine-grained KV cache and the full attention on Proof-Pile is only 0.1.

\paragraph{Weak Context-dependent Tasks}
As shown in Table~\ref{tab:res_weak},\footnote{We report the performance in which contexts are compressed at least once here. Additional results in the short-context setting can be found in Appendix~\ref{appendix: short context performance}} among four datasets, full attention shows a clear advantage only on the BBH dataset, which involves some complex reasoning tasks. In the BBH dataset, reasoning paths can usually extend over several hundred tokens. Long-form reasoning within compressed contexts frequently encounters challenges, such as generating content that spans multiple segments, which results in the accumulation of substantial inaccuracies during the process. This severely impacts the final output. However, in the other three datasets, despite the diversity of task types, the reasoning paths are typically only dozens of tokens long, which explains why compression models maintain near-lossless performance.

\begin{table}
    \centering
    \small
    \scalebox{0.80}{\begin{tabular}{c|c|cccc}
        \toprule
        Ratio & Type & MMLU-Pro & BBH & GSM8K & HellaSwag \\
        \midrule
        - & Full Attention & 34.1 & 64.8 & 51.2 & 82.8 \\
        \midrule
        \multirow{3}{*}{4} & Coarse-Rec & 34.1 & 53.8 & 50.3 & 81.9 \\
        & Coarse-KV & 35.3 & 58.1 & 48.7 & 82.3 \\
        & Fine-KV & 33.9 & 59.2 & 52.2 & 82.5 \\
        \midrule
        \multirow{3}{*}{8} & Coarse-Rec & 34.1 & 54.6 & 51.9 & 82.0 \\
        & Coarse-KV & 35.6 & 56.1 & 49.0 & 82.2 \\
        & Fine-KV & 34.6 & 56.8 & 51.9 & 82.5 \\
        \midrule
        \multirow{3}{*}{16} & Coarse-Rec & 34.1 & 53.2 & 50.0 & 81.9 \\
        & Coarse-KV & 35.6 & 55.7 & 50.1 & 82.2 \\
        & Fine-KV & 34.3 & 56.0 & 51.7 & 82.2 \\
        \midrule
        \multirow{3}{*}{32} & Coarse-Rec & 34.1 & 54.8 & 50.8 & 81.9 \\
        & Coarse-KV & 35.6 & 50.6 & 50.5 & 82.2 \\
        & Fine-KV & 33.6 & 55.0 & 50.6 & 82.2 \\
        \bottomrule
    \end{tabular}}
    \caption{Performance on weak context-dependent tasks.}
    \label{tab:res_weak}
    \vspace{-2ex}
\end{table}

\begin{table*}
    \centering
    \small
    \scalebox{0.95}{\begin{tabular}{c|c|ccccccc|c}
        \toprule
         Ratio & Compression Type & RAG & Rerank & LongQA & ICL & Synthetic & Summ. & Code & Average\\
        \midrule
        \multirow{2}{*}{-}
        & Full Attention& 61.8 & 39.9 & 41.6 & 62.3 & 93.9 & 23.8 & 66.1 & 55.6 \\
        & Full Attention, Finetune & 61.7 & 38.5 & 42.3 & 60.0 & 91.0 & 24.1 & 65.7 & 54.7 \\
        \midrule
        \multirow{3}{*}{4} & Coarse-grained, Recurrent & 49.9 & 2.1 & 35.2 & 29.4 & 11.2 & 18.2 & 59.3 & 29.3 \\
        & Coarse-grained, KV Cache & 51.7 & 5.2 & 33.9 & 36.0 & 14.2 & 17.6 & 57.8 & 30.9 \\
        & Fine-grained, KV Cache & \textbf{60.6} & \textbf{23.4} & \textbf{40.3} & \textbf{70.6} & \textbf{40.6} & \textbf{21.0} & \textbf{63.0} & \textbf{46.2} \\
        \midrule
        \multirow{3}{*}{8} & Coarse-grained, Recurrent & 49.8 & 1.3 & 36.0 & 25.9 & 11.2 & \textbf{17.7} & 58.6 & 28.6 \\
        & Coarse-grained, KV Cache & 50.8 & 3.8 & 36.5 & 33.6 & 13.5 & 16.1 & 57.2 & 30.2 \\
        & Fine-grained, KV Cache & \textbf{57.6} & \textbf{14.5} & \textbf{40.2} & \textbf{68.1} & \textbf{26.9} & 16.7 & \textbf{60.7} & \textbf{40.7} \\
        \midrule
        \multirow{3}{*}{16} & Coarse-grained, Recurrent & 49.9 & 1.4 & 34.9 & 20.8 & 11.2 & \textbf{17.8} & 57.5 & 27.6 \\
        & Coarse-grained, KV Cache & 50.2 & 4.4 & 34.2 & 29.1 & 13.1 & 16.7 & 58.1 & 29.4 \\
        & Fine-grained, KV Cache & \textbf{55.4} & \textbf{10.0} & \textbf{40.4} & \textbf{49.3} & \textbf{13.8} & \textbf{16.3} & \textbf{59.2} & \textbf{34.9} \\
        \midrule
        \multirow{3}{*}{32} & Coarse-grained, Recurrent & 49.3 & 1.2 & 33.6 & 21.1 & 11.1 & \textbf{17.5} & 58.2 & 27.4 \\
        & Coarse-grained, KV Cache & 49.9 & 2.6 & 34.2 & 25.0 & \textbf{12.2} & 17.1 & 58.2 & 28.5 \\
        & Fine-grained, KV Cache & \textbf{53.1} & \textbf{3.1} & \textbf{37.6} & \textbf{36.4} & 11.9 & 16.1 & \textbf{59.2} & \textbf{31.0} \\
        \bottomrule
    \end{tabular}}
    \caption{Performance comparison among full attention and compression architectures on long context tasks. \textbf{Bold} indicates the best result along the same compression ratio.}
    \label{tab:res_long}
    \vspace{-2ex}
\end{table*}

\paragraph{Long Context Tasks} Table~\ref{tab:res_long} presents the results, where we have the following findings: \textbf{(1) Higher Compression Ratio Leads to Lower Performance.} While Fine-KV can achieve comparable performance to full attention in some tasks at lower compression ratios (e.g., 4), it struggle to maintain this level of performance at higher ratios. \textbf{(2) The extent of performance degradation in compressed models varies significantly across different types of tasks.} For tasks where the required information is somewhat fuzzy (e.g., Summarization), or where the query is closely related to the general topics of the context (e.g., RAG and LongQA), compression does not noticeably affect the performance. For many-shot ICL, which requires almost the full context, the fine-grained KV cache can maintain performance comparable to full attention even at low compression rates. However, in tasks that demand precise rephrasing or involve highly complex multi-hop reasoning, such as Rerank\footnote{This task needs $O(n)$ to evaluate the relevance score for each candidate document, and then sort these documents with $O(n\log n)$ on average.}, none of the compressed models perform on par with full attention. \textbf{(3) Coarse-grained methods appear to struggle in fully utilizing the available memory budget.} Despite having the same memory budget, the Fine-KV's performance decreases systematically as the compression rate increases, whereas coarse-grained methods show consistently poor performance across different ratios. The trends observed in perplexity evaluation support this finding, suggesting that coarse-grained gist placement is less effective at learning how to optimize the memory budget for compression.

\section{Understanding Why and How Compression Fails}
\label{sec: understand why and how}
Previous results show that gist token-based context compression exhibits a discernible performance gap compared to full attention, particularly in tasks like synthetic recall that require exact rehearsal. This suggests the presence of a ``compression bottleneck'' that prevents the language model from treating gist tokens as equivalent to uncompressed context. We conduct a probing experiment to investigate the nature of this bottleneck and examine three critical failure modes arising from it.

\subsection{Compression Bottleneck Probing}
\paragraph{Experimental Setting} We adopt the concept of autoencoder to investigate the quality of compressed representations in gist tokens. For this experiment, we use the Fine-KV architecture, which is the most effective compression architecture according to previous results. We evaluate whether each gist token completely stores the contextual information of its corresponding snippet by training a probing decoder to recover the corresponding token sequence. We examine two decoders: an \textsc{Llama3-8B} model that inherits the full pre-trained parameters and a model with only a single transformer block. This allows us to explore the compression quality from the perspective of decoder capacities. 

\paragraph{Results}
In Table~\ref{table: probing}, we report the training loss after 2K training steps for two models, along with their token-level reconstruction accuracy on the PG19 dataset. Although the full model demonstrates superior performance, it still exhibits significant shortcomings in decoding the information within gist tokens. Under high compression ratios, the model's accuracy even falls below 20\%, indicating that it can only retain fuzzy content rather than remember the precise details from the original context. Ideally, copying a small set of recent tokens should be an easy task, yet probing experiments reveal poor performance. This suggests that the representations of current gist token memory impose a severe compression bottleneck, limiting the model’s capacity to extract and utilize contextual information effectively.

\subsection{Failure Pattern Observations}
\begin{figure*}[!t]
	\centering
	\includegraphics[width=\linewidth]{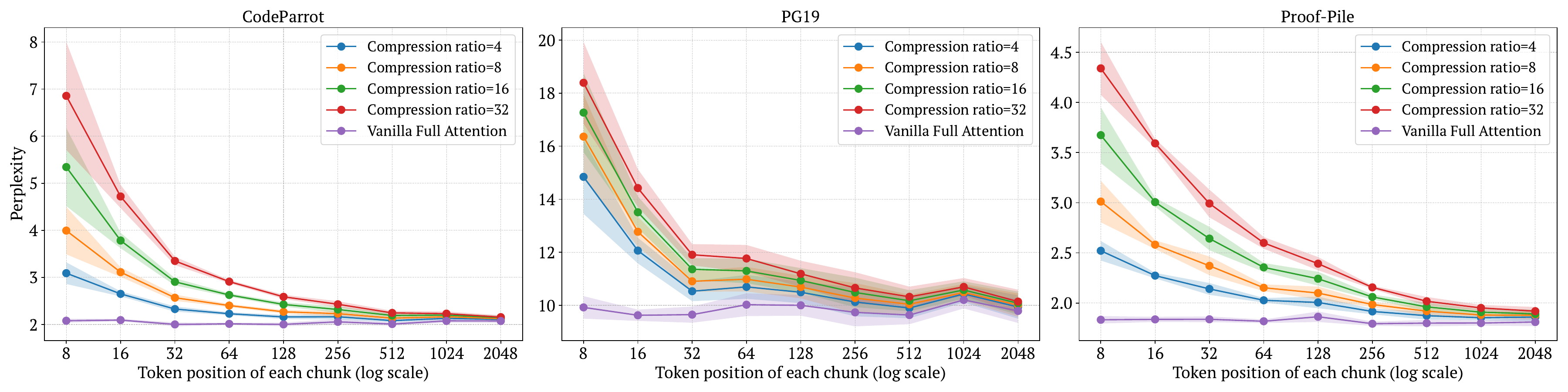}
	\caption{Average Perplexity of tokens in different positions among segments.}
	\label{fig: boundary ppl}
    \vspace{-2ex}
\end{figure*}

\begin{figure}[!t]
	\centering
	\includegraphics[width=0.95\linewidth]{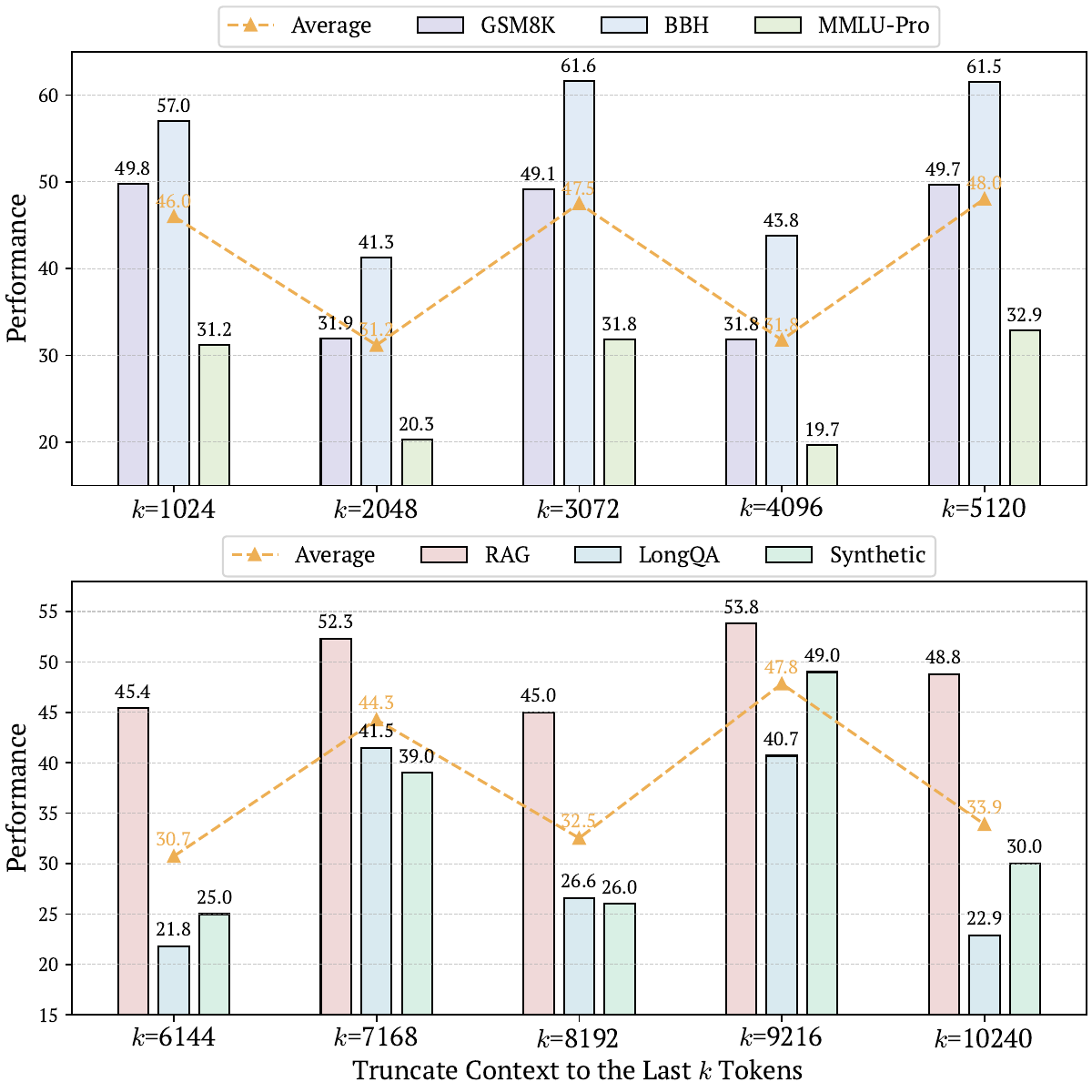}
	\caption{Performance on different tasks while truncating context to the last $k$ tokens. When $k$ is a multiple of 2048, the model will generate near the boundary.} 
	\label{fig: boundary short long}
    \vspace{-2ex}
\end{figure}

\begin{table}
    \centering
    \small
    \scalebox{0.85}{\begin{tabular}{c|c|cccc}
        \toprule
        \multirow{2}{*}{Decoder Type} & \multirow{2}{*}{Train Loss} & \multicolumn{4}{c}{Reconstruction Accuracy} \\ 
        & & 4 & 8 & 16 & 32 \\
        \midrule
        Weak & 2.64 & 53.9\% & 19.2\% & 9.6\% & 5.1\% \\
        Strong & 2.01 & 77.3\% & 39.9\% & 19.3\% & 10.0\% \\
        \bottomrule
    \end{tabular}}
    \caption{Reconstruction accuracies with different compression ratios (CR).}
    \label{table: probing}
    \vspace{-4ex}
\end{table}

\begin{figure}[!t]
	\centering
	\includegraphics[width=\linewidth]{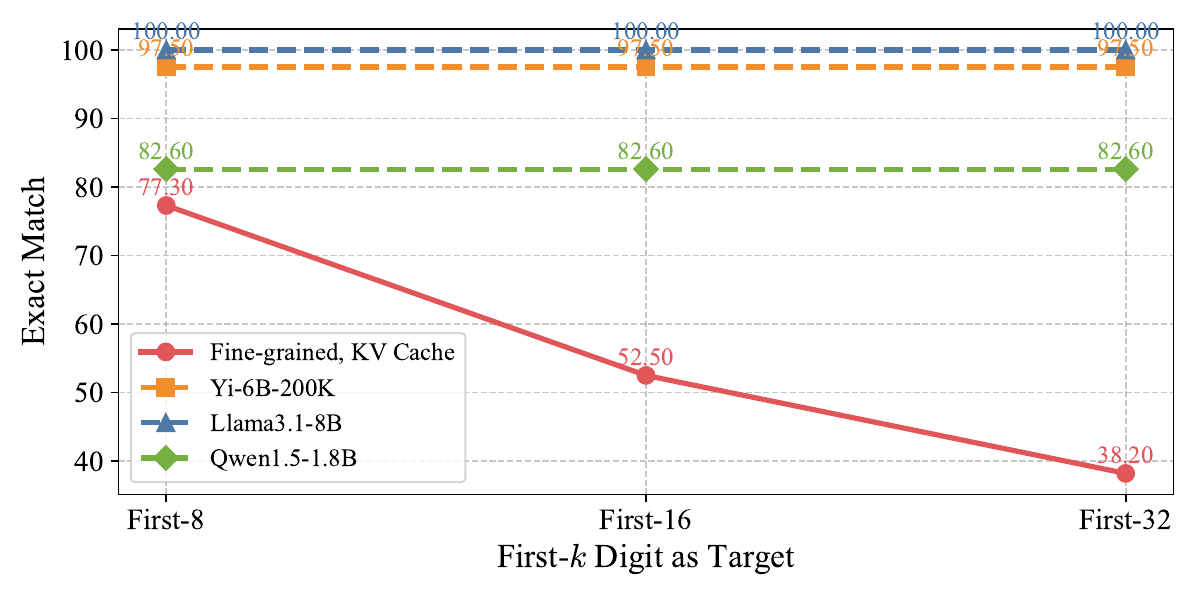}
	\caption{Performance on the 32-digit uuid recall task. We report the exact match rates of various first-$k$ digits.}
	\label{fig: Lost along the way}
    \vspace{-3ex}
\end{figure}

The compression bottleneck may evolve into specific failure patterns. We highlight three representative and interesting patterns:

\paragraph{Lost by the boundary} This discovery stems from an analysis of token-level perplexity distribution. As illustrated in Figure~\ref{fig: boundary ppl}, we compute the average perplexity of the tokens at each position within individual segments, excluding the first segment since it lacks gist tokens as contextual input. The results reveal that, while token perplexity in the full attention model remains relatively uniform across positions, the compressed model exhibits a clear pattern of higher perplexity at the start of the segment and lower perplexity toward the end.

Furthermore, we evaluated the impact on generation tasks by truncating the context to a specific length. As shown in Figure~\ref{fig: boundary short long}, with a segment length set to 2K, the performance when generation starts at the beginning of a segment is substantially worse compared to the case when generation starts from the middle of a segment. This indicates that the segment boundary effects influence not only the accuracy of reading specific information but also the model's overall language modeling capability.

\paragraph{Lost if surprise} We find that under constrained memory budgets, the model tends to prioritize retaining detailed information that closely aligns with the overarching theme of the context. To validate this, we construct a synthetic dataset\footnote{We provide an example for clarity in Table~\ref{table: synthetic popqa example}} with different configurations based on the PopQA dataset from the RAG task, as it provides explicit question subjects, and most documents are typically related to the same subject. We randomly insert a ``needle'' between sentences in the gold document, formatted as: ``\{subj\}'s special \{needle\_type\} is \{needle\_content\}''. Here, \{subj\} can either be the original subject or ``Mr. Tree'', while \{needle\_type\} can be either ``food'' or an 8-digit number. When \{subj\} is the original subject, we consider the needle to be relevant to the theme of most of the context; otherwise, it is surprising and unrelated. All needles are transformed into compressed gist tokens during the model's decoding stage. As shown in Table~\ref{table: lost if surprise}, our experimental results reveal significant performance differences in both needle types when altering only the subject of a single sentence. This indicates that the successful retrieval of compressed information is associated with its relevance to the context. An ``unexpected'' information is more likely to be lost during compression.

\begin{table}
    \centering
    \small
    \scalebox{0.85}{\begin{tabular}{c|c|cccc}
        \toprule
        \multirow{2}{*}{Needle Type} & \multirow{2}{*}{Rel.} & \multicolumn{4}{c}{Compression Ratio} \\ 
        & & 4 & 8 & 16 & 32 \\
        \midrule
        \multirow{2}{*}{Word} & \ding{51} & 89.8$_{\text{\color{gray}(+0.0)}}$ & 50.7$_{\text{\color{gray}(+0.0)}}$ & 26.0$_{\text{\color{gray}(+0.0)}}$ & 19.6$_{\text{\color{gray}(+0.0)}}$ \\
        & \ding{55} & 89.6$_{\text{\color{red}(-0.2)}}$ & 35.8$_{\text{\color{red}(-14.9)}}$ & 18.0$_{\text{\color{red}(-8.0)}}$ & 16.8$_{\text{\color{red}(-2.8)}}$ \\
        \midrule
        \multirow{2}{*}{Number} & \ding{51} & 84.5$_{\text{\color{gray}(+0.0)}}$ & 69.2$_{\text{\color{gray}(+0.0)}}$ & 26.3$_{\text{\color{gray}(+0.0)}}$ & 17.2$_{\text{\color{gray}(+0.0)}}$ \\
        & \ding{55} & 84.4$_{\text{\color{red}(-0.1)}}$ & 59.0$_{\text{\color{red}(-10.2)}}$ & 20.9$_{\text{\color{red}(-5.7)}}$ & 16.6$_{\text{\color{red}(-0.6)}}$ \\
        \bottomrule
    \end{tabular}}
    \caption{Performance on synthetic recall task (PopQA).}
    \label{table: lost if surprise}
    \vspace{-4ex}
\end{table}

\paragraph{Lost along the way} We notice that compression-enhanced architectures struggle to recover exact rehearsal effectively. When dealing with a relatively long ``needle'', the compression process can scatter critical information across multiple gist tokens. Consequently, even if the model identifies the beginning of the target information, it risks losing track during subsequent steps of generation. To validate this observation, we conducted a recall experiment using 32-digit UUIDs, comparing the performance of full attention models against compressed models, and analyzed their accuracy across prefixes of varying lengths. As illustrated in Figure~\ref{fig: Lost along the way}, the replication accuracy of full attention models remains stable regardless of prefix length, suggesting that once the starting point is identified, copying the rest of the content is straightforward. In contrast, compressed models show a significant drop in accuracy, decreasing to less than half of the original as the prefix extends from the first four digits to all 32 digits. This finding highlights the reduced copying reliability associated with compressed representations.

\begin{table*}
    \centering
    \small
    \scalebox{0.9}{\begin{tabular}{c|c|ccccccc|c}
        \toprule
         Ratio & Compression Type & RAG & Rerank & LongQA & ICL & Synthetic & Summ. & Code & Average\\
        \midrule
        -
        & Full Attention& 61.8 & 39.9 & 41.6 & 62.3 & 93.9 & 23.8 & 66.1 & 55.6 \\
        \midrule
        \multirow{4}{*}{4} & Fine-grained, KV Cache & 60.6$_\text{\color{black}(+0.0)}$ & 23.4$_\text{\color{black}(+0.0)}$ & 40.3$_\text{\color{black}(+0.0)}$ & 70.6$_\text{\color{black}(+0.0)}$ & 40.6$_\text{\color{black}(+0.0)}$ & 21.0$_\text{\color{black}(+0.0)}$ & 62.0$_\text{\color{black}(+0.0)}$ & 46.1$_\text{\color{black}(+0.0)}$ \\
        & + Fine-grained AE & 60.9$_\text{\color{red}(+0.3)}$ & 27.4$_\text{\color{red}(+4.0)}$ & 40.8$_\text{\color{red}(+0.5)}$ & 72.0$_\text{\color{red}(+1.4)}$ & 62.0$_\text{\color{red}(+21.4)}$ & 22.3$_\text{\color{red}(+1.3)}$ & 62.9$_\text{\color{red}(+0.9)}$ & 49.8$_\text{\color{red}(+3.7)}$ \\
        & + Segment-wise TIE & 60.4$_\text{\color{blue}(-0.2)}$ & 27.0$_\text{\color{red}(+3.6)}$ & 41.2$_\text{\color{red}(+0.9)}$ & 72.7$_\text{\color{red}(+2.1)}$ & 54.3$_\text{\color{red}(+13.7)}$ & 20.2$_\text{\color{blue}(-0.8)}$ & 62.1$_\text{\color{red}(+0.1)}$ & 48.3$_\text{\color{red}(+2.2)}$ \\
        & + Both Strategies & 61.1$_\text{\color{red}(+0.5)}$ & 27.4$_\text{\color{red}(+4.0)}$ & 40.3$_\text{\color{red}(+0.0)}$ & 75.0$_\text{\color{red}(+4.4)}$ & 62.1$_\text{\color{red}(+21.5)}$ & 22.2$_\text{\color{red}(+1.2)}$ & 62.9$_\text{\color{red}(+0.9)}$ & \textbf{50.1}$_\text{\color{red}(+4.0)}$ \\
        
        \midrule
        
        \multirow{4}{*}{8} & Fine-grained, KV Cache & 57.6$_\text{\color{black}(+0.0)}$ & 14.5$_\text{\color{black}(+0.0)}$ & 40.2$_\text{\color{black}(+0.0)}$ & 68.1$_\text{\color{black}(+0.0)}$ & 26.9$_\text{\color{black}(+0.0)}$ & 16.7$_\text{\color{black}(+0.0)}$ & 60.7$_\text{\color{black}(+0.0)}$ & 40.7$_\text{\color{black}(+0.0)}$ \\
        & + Fine-grained AE & 58.3$_\text{\color{red}(+0.7)}$ & 15.6$_\text{\color{red}(+0.9)}$ & 39.8$_\text{\color{blue}(-0.4)}$ & 68.7$_\text{\color{red}(+0.6)}$ & 34.8$_\text{\color{red}(+7.9)}$ & 18.5$_\text{\color{red}(+1.8)}$ & 61.3$_\text{\color{red}(+0.6)}$ & 42.4$_\text{\color{red}(+1.7)}$ \\
        & + Segment-wise TIE & 58.1$_\text{\color{red}(+0.4)}$ & 17.6$_\text{\color{red}(+3.1)}$ & 40.0$_\text{\color{blue}(-0.2)}$ & 70.0$_\text{\color{red}(+1.9)}$ & 30.2$_\text{\color{red}(+3.3)}$ & 17.7$_\text{\color{red}(+1.0)}$ & 60.7$_\text{\color{red}(+0.0)}$ & 42.0$_\text{\color{red}(+1.3)}$ \\
        & + Both Strategies & 58.3$_\text{\color{red}(+0.7)}$ & 19.7$_\text{\color{red}(+5.2)}$ & 40.4$_\text{\color{red}(+0.0)}$ & 70.7$_\text{\color{red}(+2.6)}$ & 35.2$_\text{\color{red}(+8.9)}$ & 19.5$_\text{\color{red}(+2.8)}$ & 61.4$_\text{\color{red}(+0.7)}$ & \textbf{43.6}$_\text{\color{red}(+2.9)}$ \\
        
        \midrule
        
        \multirow{4}{*}{16} & Fine-grained, KV Cache & 55.4$_\text{\color{black}(+0.0)}$ & 10.0$_\text{\color{black}(+0.0)}$ & 40.4$_\text{\color{black}(+0.0)}$ & 49.3$_\text{\color{black}(+0.0)}$ & 13.8$_\text{\color{black}(+0.0)}$ & 16.3$_\text{\color{black}(+0.0)}$ & 59.2$_\text{\color{black}(+0.0)}$ & 34.9$_\text{\color{black}(+0.0)}$ \\
        & + Fine-grained AE & 55.6$_\text{\color{red}(+0.2)}$ & 11.3$_\text{\color{red}(+1.3)}$ & 40.4$_\text{\color{red}(+0.0)}$ & 47.1$_\text{\color{red}(+0.3)}$ & 14.7$_\text{\color{red}(+0.9)}$ & 16.2$_\text{\color{blue}(-0.1)}$ & 59.6$_\text{\color{red}(+0.4)}$ & 35.0$_\text{\color{red}(+0.1)}$ \\
        & + Segment-wise TIE & 55.6$_\text{\color{red}(+0.2)}$ & 10.4$_\text{\color{red}(+0.4)}$ & 40.7$_\text{\color{red}(+0.3)}$ & 55.5$_\text{\color{red}(+8.4)}$ & 14.8$_\text{\color{red}(+1.0)}$ & 15.3$_\text{\color{blue}(-1.0)}$ & 58.1$_\text{\color{blue}(-1.1)}$ & 35.7$_\text{\color{red}(+0.8)}$ \\
        & + Both Strategies & 56.3$_\text{\color{red}(+0.9)}$ & 12.7$_\text{\color{red}(+2.7)}$ & 41.7$_\text{\color{red}(+1.3)}$ & 56.3$_\text{\color{red}(+7.0)}$ & 14.9$_\text{\color{red}(+1.1)}$ & 15.7$_\text{\color{blue}(-0.6)}$ & 59.6$_\text{\color{red}(+0.4)}$ & \textbf{36.7}$_\text{\color{red}(+1.8)}$ \\
        \midrule
        \multirow{4}{*}{32} & Fine-grained, KV Cache & 53.1$_\text{\color{black}(+0.0)}$ & 3.1$_\text{\color{black}(+0.0)}$ & 37.6$_\text{\color{black}(+0.0)}$ & 36.4$_\text{\color{black}(+0.0)}$ & 11.9$_\text{\color{black}(+0.0)}$ & 16.1$_\text{\color{black}(+0.0)}$ & 59.2$_\text{\color{black}(+0.0)}$ & 31.0$_\text{\color{black}(+0.0)}$ \\
        & + Fine-grained AE & 54.3$_\text{\color{red}(+1.2)}$ & 4.6$_\text{\color{red}(+1.5)}$ & 39.3$_\text{\color{red}(+1.7)}$ & 34.1$_\text{\color{blue}(-2.3)}$ & 13.1$_\text{\color{red}(+1.2)}$ & 17.1$_\text{\color{red}(+1.0)}$ & 59.8$_\text{\color{red}(+0.6)}$ & 31.8$_\text{\color{red}(+0.8)}$ \\
        & + Segment-wise TIE & 53.1$_\text{\color{red}(+0.0)}$ & 4.6$_\text{\color{red}(+1.5)}$ & 40.3$_\text{\color{red}(+2.7)}$ & 43.6$_\text{\color{red}(+7.2)}$ & 13.1$_\text{\color{red}(+1.2)}$ & 17.0$_\text{\color{red}(+0.9)}$ & 59.8$_\text{\color{red}(+0.6)}$ & \textbf{33.1}$_\text{\color{red}(+2.1)}$ \\
        & + Both Strategies & 54.4$_\text{\color{red}(+1.3)}$ & 4.9$_\text{\color{red}(+1.8)}$ & 39.8$_\text{\color{red}(+2.2)}$ & 41.8$_\text{\color{red}(+5.4)}$ & 13.1$_\text{\color{red}(+0.9)}$ & 17.1$_\text{\color{red}(+1.0)}$ & 59.8$_\text{\color{red}(+0.6)}$ & 33.0$_\text{\color{red}(+2.0)}$ \\
        \bottomrule
    \end{tabular}}
    \caption{
Performance comparisons using our methods, with the best ``average'' results bolded for clarity.}
    \label{table: long context improvements}
    \vspace{-2ex}
\end{table*}

\section{Mitigating Compression Flaws}
\label{sec:mitigating}

\subsection{Methodology}

Building on these findings, we have identified critical shortcomings in the current architecture's context compression. In this section, we propose two effective learning strategies to address them.

\paragraph{Fine-grained Autoencoding (AE)} The probing experiments in Section~\ref{sec: understand why and how} indicate that the compressed representations of current gist tokens struggle to reconstruct the original content. To address this issue, we introduce an additional autoencoding loss during training to explicitly encourage the retention of the original contextual information. Different from ICAE~\cite{ICAE}, we require each gist token to be responsible for a specific snippet. Following the mainstream conclusion in autoencoding research that weak decoders help learn better representations~\cite{LessIsMore}, we adopt a single-layer transformer as the decoder. For each gist token $g_i^{kv}$, the objective is to reconstruct the original token sequence between the current and previous gist tokens. The input for this task is:
\begin{align*}
    \lbrack g_i^{kv}, {\lbrack \text{ae} \rbrack}_{r}, x_1, \ldots, x_{r}\rbrack
\end{align*}
where ${\lbrack \text{ae} \rbrack}_{r}$ is a special token to prompt model to reconstruct $r$ tokens (i.e., $x_1$ to $x_r$). The loss of autoencoding is similarly defined in an auto-regressive way:
\begin{align*}
    \mathcal{L}_{\text{ae}} = \frac{1}{N} \frac{1}{r} \sum^N_{i=1} \sum^r_{j=1} \log P_\theta(x_j | g_i^{kv}, {\lbrack \text{ae} \rbrack}_{r}, x_{<j})
\end{align*}

\paragraph{Segment-wise Token Importance Estimation (TIE)}
Another approach to promote compression is to adjust the loss weights of different tokens, since each token depends on the context in different degrees. We hypothesize that the importance of a token is determined by the modeling difficulty it presents during segment-wise compression. The more a token relies on the compressed gist context for prediction, the more effort should be dedicated to learning it. Inspired by LongPPL~\cite{LongPPL}, we estimate the reliance of each token ($x_i$) on the gist context and allocate a tailored learning weight $w_i$ accordingly: 
\begin{align*}
    \text{Diff}(x_i) &= \text{min} (\log \frac{P_\theta (x_i | x_{<i}^{\text{seg}})}{P_\theta(x_i | x_{<i}^{\text{full}})}, \gamma), \\
    w_i &= \frac{e^{\text{Diff}(x_i)}}{\sum^N_{j=1}e^{\text{Diff}(x_j)}}.
\end{align*}
Here, $P_\theta$ denotes the original language model, $x_{<i}^{\text{seg}}$ denotes the preceding tokens only in the current segment, and $x_{<i}^{\text{full}}$ denotes the full context, including tokens in previous segments. This reliance is quantified by analyzing the difference in modeling probabilities when the token attends to the full context versus the local segment alone.

\subsection{Experiments}

\paragraph{Boundary Effect Test}
\begin{table}
    \centering
    \small
    \scalebox{0.85}{\begin{tabular}{c|c|ccc}
        \toprule
        $k$ & Model & MMLU-Pro & BBH & GSM8K \\ 
        \midrule
        \multirow{3}{*}{2048} & Fine-grained KV & 20.3$_{\text{\color{gray}(+0.0)}}$ & 41.3$_{\text{\color{gray}(+0.0)}}$ & 31.9$_{\text{\color{gray}(+0.0)}}$ \\
        & + Fine-grained AE & 23.4$_{\text{\color{red}(+3.1)}}$ & 47.8$_{\text{\color{red}(+6.5)}}$ & 34.3$_{\text{\color{red}(+2.4)}}$ \\
        & + Segment-wise TIE & 22.9$_{\text{\color{red}(+2.6)}}$ & 46.3$_{\text{\color{red}(+5.0)}}$ & 32.3$_{\text{\color{red}(+2.0)}}$ \\
        \midrule
        \multirow{3}{*}{4096} & Fine-grained KV & 19.7$_{\text{\color{gray}(+0.0)}}$ & 43.8$_{\text{\color{gray}(+0.0)}}$ & 31.8$_{\text{\color{gray}(+0.0)}}$ \\
        & + Fine-grained AE & 22.5$_{\text{\color{red}(+2.8)}}$ & 51.0$_{\text{\color{red}(+7.2)}}$ & 35.1$_{\text{\color{red}(+3.3)}}$ \\
        & + Segment-wise TIE & 22.9$_{\text{\color{red}(+3.2)}}$ & 50.8$_{\text{\color{red}(+7.0)}}$ & 34.7$_{\text{\color{red}(+2.9)}}$ \\
        \bottomrule
    \end{tabular}}
        \caption{Improvements of our mitigating methods on the ``lost by the boundary'' problem.}
    \label{table: boundary improvements}
    \vspace{-2ex}
\end{table}

Previous results show that gist-based models demonstrate strong performance on weak context-dependent tasks but are severely constrained by the ``lost by the boundary'' phenomenon. We test two improved methods under the same experimental conditions in Section~\ref{sec: understand why and how}, with the results presented in Table~\ref{table: boundary improvements}. Both methods significantly enhance performance in boundary regions, particularly on the BBH dataset, which involves tasks requiring long-form reasoning. This improvement may be attributed to their ability to reduce the accumulation of errors during the generation process. While these methods do not completely eliminate the boundary effect, they offer promising strategies for mitigating its impact.

\paragraph{Long Context Tasks}
Table~\ref{table: long context improvements} highlights that both methods consistently enhance the model's performance on long-context tasks, particularly under low compression ratios. Key observations include: (1) For tasks where the performance gap between the compression-enhanced model and full attention is relatively small (e.g., RAG and LongQA), both methods maintain excellent performance without negative impacts. For the many-shot ICL task, they even demonstrate continuous improvements. (2) For tasks where the original architectures struggle, such as rerank and synthetic recall, both methods deliver remarkable performance gains. For instance, under a compression ratio of 4, the improvements on the synthetic recall task reach as high as 52.7\% and 33.7\%, respectively. These indicate that our methods can effectively enhance the model to read context information from gist tokens.

\section{Related Work}

\paragraph{KV Cache Compression}
Recent work has explored KV cache optimization at the \textit{layer}, \textit{head}, \textit{token}, and \textit{tensor} levels. \textit{Layer}-level methods merge caches across layers using inter-layer similarities~\cite{CLA, YOCO, LayerCondensedKVCache, MiniCache}. \textit{Head}-level techniques allow multiple query heads to share key-value pairs~\cite{GQA, MQA}. \textit{Tensor}-level approaches, such as low-rank approximations, compress caches into compact representations~\citep{Deepseek-v2}, while quantization reduces precision for memory savings~\cite{KIVI}. \textit{Token}-level methods preserve only critical tokens, including learnable tokens~\cite{Gist, ICAE, Nugget, Landmark, AutoCompressors, Beacon}, token eviction~\cite{H2O, Scissorhands, FastGen}, external memory~\cite{InfLLM}, and hard selection~\cite{selective-tokens, LLMLingua}. In this work, we focus on the direction that introduces a few learnable special tokens to replace the previous full context.

\paragraph{Sparse Attention}
Researchers have been exploring efficient alternatives of full attention~\cite{longformer, bigbird, reformer, socialformer, EfficientTransformer-Survey}. 
Recently, it has been widely observed that LLMs naturally exhibit significant sparse attention patterns, especially in long-form texts~\cite{MInference}. To leverage such characteristics, researchers have developed heuristic or learnable sparsification strategies that achieve significant speedup while maintaining reliable performance~\cite{MInference, DuoAttention}. The gist token-based context compression approach can be regarded as a special case of sparse attention with a segment-wise approach~\cite{AutoCompressors, Beacon}: where full attention is employed within each segment.

\section{Conclusion}
Our comprehensive evaluation presents that while gist-based context compression shows promise as an alternative to full attention in many tasks, it still falls short in specific scenarios. Through carefully designed probing experiments, we identify critical compression bottlenecks and typical failure modes. Furthermore, we propose two effective strategies that significantly enhance compression performance. These findings offer new insights and directions for advancing context compression techniques in the future.

\clearpage
\section*{Limitations}
\paragraph{Model Scale and Context Length} Constrained by our available computational resource, we are able to train long-text large language models with sizes up to 7/8B parameters in a 16K context window. Larger models (e.g., Llama3.1-70B) typically have more layers, which enables them to offer greater memory capacity and stronger reading capabilities under the same compression ratio when using gist token-based compression. Thus, such larger models may offer advantages in reducing performance degradation, but this still needs to be verified in future studies.

\paragraph{Scope of Compression Methods}
Our study concentrates on a comparative analysis between gist token-based context compression and the full attention mechanism. While other techniques, such as token-dropping methods represented by StreamingLLM and H2O, are also capable of context compression, including them in our scope would go beyond the focus of this paper. Our primary aim is to investigate the effectiveness and limitations of gist token-based context compression, using full attention as the ideal performance upper bound for comparison. Incorporating additional methods would risk complicating the analysis and diluting the focus on the central research question. Therefore, we choose to maintain the scope to ensure clarity and depth in our insights and analysis.

\section*{Ethical Discussion}
This study focuses on the performance of gist token-based context compression techniques, without introducing explicitly designed features that could directly influence the cognition of language models. We select widely recognized and validated public training datasets. This can minimize the risk of injecting new biases or toxic data. These datasets are typically subjected to rigorous review and curation, ensuring balanced and stable data distributions. As a result, they help mitigate the impact of harmful information on the model’s learning process and prevent significant distortions in its cognitive and decision-making patterns.


\begin{thebibliography}{56}
\providecommand{\natexlab}[1]{#1}

\bibitem[{Ainslie et~al.(2023)Ainslie, Lee{-}Thorp, de~Jong, Zemlyanskiy, Lebr{\'{o}}n, and Sanghai}]{GQA}
Joshua Ainslie, James Lee{-}Thorp, Michiel de~Jong, Yury Zemlyanskiy, Federico Lebr{\'{o}}n, and Sumit Sanghai. 2023.
\newblock \href {https://doi.org/10.18653/V1/2023.EMNLP-MAIN.298} {{GQA:} training generalized multi-query transformer models from multi-head checkpoints}.
\newblock In \emph{Proceedings of the 2023 Conference on Empirical Methods in Natural Language Processing, {EMNLP} 2023, Singapore, December 6-10, 2023}, pages 4895--4901. Association for Computational Linguistics.

\bibitem[{Beltagy et~al.(2020)Beltagy, Peters, and Cohan}]{longformer}
Iz~Beltagy, Matthew~E. Peters, and Arman Cohan. 2020.
\newblock \href {https://arxiv.org/abs/2004.05150} {Longformer: The long-document transformer}.
\newblock \emph{CoRR}, abs/2004.05150.

\bibitem[{Brandon et~al.(2024)Brandon, Mishra, Nrusimha, Panda, and Ragan{-}Kelley}]{CLA}
William Brandon, Mayank Mishra, Aniruddha Nrusimha, Rameswar Panda, and Jonathan Ragan{-}Kelley. 2024.
\newblock \href {https://doi.org/10.48550/ARXIV.2405.12981} {Reducing transformer key-value cache size with cross-layer attention}.
\newblock \emph{CoRR}, abs/2405.12981.

\bibitem[{Bulatov et~al.(2022)Bulatov, Kuratov, and Burtsev}]{RMT}
Aydar Bulatov, Yuri Kuratov, and Mikhail~S. Burtsev. 2022.
\newblock \href {https://doi.org/10.48550/ARXIV.2207.06881} {Recurrent memory transformer}.
\newblock \emph{CoRR}, abs/2207.06881.

\bibitem[{Chen et~al.(2023)Chen, Wong, Chen, and Tian}]{PI}
Shouyuan Chen, Sherman Wong, Liangjian Chen, and Yuandong Tian. 2023.
\newblock \href {https://doi.org/10.48550/ARXIV.2306.15595} {Extending context window of large language models via positional interpolation}.
\newblock \emph{CoRR}, abs/2306.15595.

\bibitem[{Chen et~al.(2024)Chen, Qian, Tang, Lai, Liu, Han, and Jia}]{LongLoRA}
Yukang Chen, Shengju Qian, Haotian Tang, Xin Lai, Zhijian Liu, Song Han, and Jiaya Jia. 2024.
\newblock \href {https://openreview.net/forum?id=6PmJoRfdaK} {Longlora: Efficient fine-tuning of long-context large language models}.
\newblock In \emph{The Twelfth International Conference on Learning Representations, {ICLR} 2024, Vienna, Austria, May 7-11, 2024}. OpenReview.net.

\bibitem[{Chevalier et~al.(2023)Chevalier, Wettig, Ajith, and Chen}]{AutoCompressors}
Alexis Chevalier, Alexander Wettig, Anirudh Ajith, and Danqi Chen. 2023.
\newblock \href {https://doi.org/10.18653/V1/2023.EMNLP-MAIN.232} {Adapting language models to compress contexts}.
\newblock In \emph{Proceedings of the 2023 Conference on Empirical Methods in Natural Language Processing, {EMNLP} 2023, Singapore, December 6-10, 2023}, pages 3829--3846. Association for Computational Linguistics.

\bibitem[{Cobbe et~al.(2021)Cobbe, Kosaraju, Bavarian, Chen, Jun, Kaiser, Plappert, Tworek, Hilton, Nakano, Hesse, and Schulman}]{GSM8K}
Karl Cobbe, Vineet Kosaraju, Mohammad Bavarian, Mark Chen, Heewoo Jun, Lukasz Kaiser, Matthias Plappert, Jerry Tworek, Jacob Hilton, Reiichiro Nakano, Christopher Hesse, and John Schulman. 2021.
\newblock \href {https://arxiv.org/abs/2110.14168} {Training verifiers to solve math word problems}.
\newblock \emph{CoRR}, abs/2110.14168.

\bibitem[{CodeParrot()}]{CodeParrot}
CodeParrot.
\newblock \href {https://huggingface.co/codeparrot/codeparrot} {https://huggingface.co/codeparrot/codeparrot}.

\bibitem[{DeepSeek{-}AI(2024)}]{Deepseek-v2}
DeepSeek{-}AI. 2024.
\newblock \href {https://doi.org/10.48550/ARXIV.2405.04434} {Deepseek-v2: {A} strong, economical, and efficient mixture-of-experts language model}.
\newblock \emph{CoRR}, abs/2405.04434.

\bibitem[{Fang et~al.(2024)Fang, Wang, Liu, Zhang, Jegelka, Gao, Ding, and Wang}]{LongPPL}
Lizhe Fang, Yifei Wang, Zhaoyang Liu, Chenheng Zhang, Stefanie Jegelka, Jinyang Gao, Bolin Ding, and Yisen Wang. 2024.
\newblock \href {https://doi.org/10.48550/ARXIV.2410.23771} {What is wrong with perplexity for long-context language modeling?}
\newblock \emph{CoRR}, abs/2410.23771.

\bibitem[{Fu et~al.(2024)Fu, Panda, Niu, Yue, Hajishirzi, Kim, and Peng}]{LongContextDataEngineering}
Yao Fu, Rameswar Panda, Xinyao Niu, Xiang Yue, Hannaneh Hajishirzi, Yoon Kim, and Hao Peng. 2024.
\newblock \href {https://openreview.net/forum?id=TaAqeo7lUh} {Data engineering for scaling language models to 128k context}.
\newblock In \emph{Forty-first International Conference on Machine Learning, {ICML} 2024, Vienna, Austria, July 21-27, 2024}. OpenReview.net.

\bibitem[{Gao et~al.(2024)Gao, Wettig, Yen, and Chen}]{Prolong}
Tianyu Gao, Alexander Wettig, Howard Yen, and Danqi Chen. 2024.
\newblock \href {https://doi.org/10.48550/ARXIV.2410.02660} {How to train long-context language models (effectively)}.
\newblock \emph{CoRR}, abs/2410.02660.

\bibitem[{Gao et~al.(2023)Gao, Xiong, Gao, Jia, Pan, Bi, Dai, Sun, Guo, Wang, and Wang}]{rag-survey}
Yunfan Gao, Yun Xiong, Xinyu Gao, Kangxiang Jia, Jinliu Pan, Yuxi Bi, Yi~Dai, Jiawei Sun, Qianyu Guo, Meng Wang, and Haofen Wang. 2023.
\newblock \href {https://doi.org/10.48550/ARXIV.2312.10997} {Retrieval-augmented generation for large language models: {A} survey}.
\newblock \emph{CoRR}, abs/2312.10997.

\bibitem[{Ge et~al.(2024{\natexlab{a}})Ge, Zhang, Liu, Zhang, Han, and Gao}]{FastGen}
Suyu Ge, Yunan Zhang, Liyuan Liu, Minjia Zhang, Jiawei Han, and Jianfeng Gao. 2024{\natexlab{a}}.
\newblock \href {https://openreview.net/forum?id=uNrFpDPMyo} {Model tells you what to discard: Adaptive {KV} cache compression for llms}.
\newblock In \emph{The Twelfth International Conference on Learning Representations, {ICLR} 2024, Vienna, Austria, May 7-11, 2024}. OpenReview.net.

\bibitem[{Ge et~al.(2024{\natexlab{b}})Ge, Hu, Wang, Wang, Chen, and Wei}]{ICAE}
Tao Ge, Jing Hu, Lei Wang, Xun Wang, Si{-}Qing Chen, and Furu Wei. 2024{\natexlab{b}}.
\newblock \href {https://openreview.net/forum?id=uREj4ZuGJE} {In-context autoencoder for context compression in a large language model}.
\newblock In \emph{The Twelfth International Conference on Learning Representations, {ICLR} 2024, Vienna, Austria, May 7-11, 2024}. OpenReview.net.

\bibitem[{Hsieh et~al.(2024)Hsieh, Sun, Kriman, Acharya, Rekesh, Jia, Zhang, and Ginsburg}]{RULER}
Cheng{-}Ping Hsieh, Simeng Sun, Samuel Kriman, Shantanu Acharya, Dima Rekesh, Fei Jia, Yang Zhang, and Boris Ginsburg. 2024.
\newblock \href {https://doi.org/10.48550/ARXIV.2404.06654} {{RULER:} what's the real context size of your long-context language models?}
\newblock \emph{CoRR}, abs/2404.06654.

\bibitem[{Jiang et~al.(2024{\natexlab{a}})Jiang, Li, Zhang, Wu, Luo, Ahn, Han, Abdi, Li, Lin, Yang, and Qiu}]{MInference}
Huiqiang Jiang, Yucheng Li, Chengruidong Zhang, Qianhui Wu, Xufang Luo, Surin Ahn, Zhenhua Han, Amir~H. Abdi, Dongsheng Li, Chin{-}Yew Lin, Yuqing Yang, and Lili Qiu. 2024{\natexlab{a}}.
\newblock \href {https://doi.org/10.48550/ARXIV.2407.02490} {Minference 1.0: Accelerating pre-filling for long-context llms via dynamic sparse attention}.
\newblock \emph{CoRR}, abs/2407.02490.

\bibitem[{Jiang et~al.(2024{\natexlab{b}})Jiang, Wu, Luo, Li, Lin, Yang, and Qiu}]{LLMLingua}
Huiqiang Jiang, Qianhui Wu, Xufang Luo, Dongsheng Li, Chin{-}Yew Lin, Yuqing Yang, and Lili Qiu. 2024{\natexlab{b}}.
\newblock \href {https://doi.org/10.18653/V1/2024.ACL-LONG.91} {Longllmlingua: Accelerating and enhancing llms in long context scenarios via prompt compression}.
\newblock In \emph{Proceedings of the 62nd Annual Meeting of the Association for Computational Linguistics (Volume 1: Long Papers), {ACL} 2024, Bangkok, Thailand, August 11-16, 2024}, pages 1658--1677. Association for Computational Linguistics.

\bibitem[{Kitaev et~al.(2020)Kitaev, Kaiser, and Levskaya}]{reformer}
Nikita Kitaev, Lukasz Kaiser, and Anselm Levskaya. 2020.
\newblock \href {https://openreview.net/forum?id=rkgNKkHtvB} {Reformer: The efficient transformer}.
\newblock In \emph{8th International Conference on Learning Representations, {ICLR} 2020, Addis Ababa, Ethiopia, April 26-30, 2020}. OpenReview.net.

\bibitem[{Kryscinski et~al.(2022)Kryscinski, Rajani, Agarwal, Xiong, and Radev}]{BookSum}
Wojciech Kryscinski, Nazneen Rajani, Divyansh Agarwal, Caiming Xiong, and Dragomir Radev. 2022.
\newblock \href {https://doi.org/10.18653/V1/2022.FINDINGS-EMNLP.488} {{BOOKSUM:} {A} collection of datasets for long-form narrative summarization}.
\newblock In \emph{Findings of the Association for Computational Linguistics: {EMNLP} 2022, Abu Dhabi, United Arab Emirates, December 7-11, 2022}, pages 6536--6558. Association for Computational Linguistics.

\bibitem[{Li et~al.(2023)Li, Dong, Guerin, and Lin}]{selective-tokens}
Yucheng Li, Bo~Dong, Frank Guerin, and Chenghua Lin. 2023.
\newblock \href {https://doi.org/10.18653/V1/2023.EMNLP-MAIN.391} {Compressing context to enhance inference efficiency of large language models}.
\newblock In \emph{Proceedings of the 2023 Conference on Empirical Methods in Natural Language Processing, {EMNLP} 2023, Singapore, December 6-10, 2023}, pages 6342--6353. Association for Computational Linguistics.

\bibitem[{Lightman et~al.(2024)Lightman, Kosaraju, Burda, Edwards, Baker, Lee, Leike, Schulman, Sutskever, and Cobbe}]{verify-step-by-step}
Hunter Lightman, Vineet Kosaraju, Yuri Burda, Harrison Edwards, Bowen Baker, Teddy Lee, Jan Leike, John Schulman, Ilya Sutskever, and Karl Cobbe. 2024.
\newblock \href {https://openreview.net/forum?id=v8L0pN6EOi} {Let's verify step by step}.
\newblock In \emph{The Twelfth International Conference on Learning Representations, {ICLR} 2024, Vienna, Austria, May 7-11, 2024}. OpenReview.net.

\bibitem[{Liu et~al.(2024{\natexlab{a}})Liu, Liu, Pan, He, Haffari, and Zhuang}]{MiniCache}
Akide Liu, Jing Liu, Zizheng Pan, Yefei He, Gholamreza Haffari, and Bohan Zhuang. 2024{\natexlab{a}}.
\newblock \href {https://doi.org/10.48550/ARXIV.2405.14366} {Minicache: {KV} cache compression in depth dimension for large language models}.
\newblock \emph{CoRR}, abs/2405.14366.

\bibitem[{Liu et~al.(2023)Liu, Desai, Liao, Wang, Xie, Xu, Kyrillidis, and Shrivastava}]{Scissorhands}
Zichang Liu, Aditya Desai, Fangshuo Liao, Weitao Wang, Victor Xie, Zhaozhuo Xu, Anastasios Kyrillidis, and Anshumali Shrivastava. 2023.
\newblock \href {http://papers.nips.cc/paper\_files/paper/2023/hash/a452a7c6c463e4ae8fbdc614c6e983e6-Abstract-Conference.html} {Scissorhands: Exploiting the persistence of importance hypothesis for {LLM} {KV} cache compression at test time}.
\newblock In \emph{Advances in Neural Information Processing Systems 36: Annual Conference on Neural Information Processing Systems 2023, NeurIPS 2023, New Orleans, LA, USA, December 10 - 16, 2023}.

\bibitem[{Liu et~al.(2024{\natexlab{b}})Liu, Yuan, Jin, Zhong, Xu, Braverman, Chen, and Hu}]{KIVI}
Zirui Liu, Jiayi Yuan, Hongye Jin, Shaochen Zhong, Zhaozhuo Xu, Vladimir Braverman, Beidi Chen, and Xia Hu. 2024{\natexlab{b}}.
\newblock \href {https://openreview.net/forum?id=L057s2Rq8O} {{KIVI:} {A} tuning-free asymmetric 2bit quantization for {KV} cache}.
\newblock In \emph{Forty-first International Conference on Machine Learning, {ICML} 2024, Vienna, Austria, July 21-27, 2024}. OpenReview.net.

\bibitem[{Lu et~al.(2021)Lu, He, Xiong, Ke, Malik, Dou, Bennett, Liu, and Overwijk}]{LessIsMore}
Shuqi Lu, Di~He, Chenyan Xiong, Guolin Ke, Waleed Malik, Zhicheng Dou, Paul Bennett, Tie{-}Yan Liu, and Arnold Overwijk. 2021.
\newblock \href {https://doi.org/10.18653/V1/2021.EMNLP-MAIN.220} {Less is more: Pretrain a strong siamese encoder for dense text retrieval using a weak decoder}.
\newblock In \emph{Proceedings of the 2021 Conference on Empirical Methods in Natural Language Processing, {EMNLP} 2021, Virtual Event / Punta Cana, Dominican Republic, 7-11 November, 2021}, pages 2780--2791. Association for Computational Linguistics.

\bibitem[{Meta-Llama(2024)}]{llama3}
Meta-Llama. 2024.
\newblock \href {https://doi.org/10.48550/ARXIV.2407.21783} {The llama 3 herd of models}.
\newblock \emph{CoRR}, abs/2407.21783.

\bibitem[{Mohtashami and Jaggi(2023)}]{Landmark}
Amirkeivan Mohtashami and Martin Jaggi. 2023.
\newblock \href {http://papers.nips.cc/paper\_files/paper/2023/hash/ab05dc8bf36a9f66edbff6992ec86f56-Abstract-Conference.html} {Random-access infinite context length for transformers}.
\newblock In \emph{Advances in Neural Information Processing Systems 36: Annual Conference on Neural Information Processing Systems 2023, NeurIPS 2023, New Orleans, LA, USA, December 10 - 16, 2023}.

\bibitem[{Mu et~al.(2023)Mu, Li, and Goodman}]{Gist}
Jesse Mu, Xiang Li, and Noah~D. Goodman. 2023.
\newblock \href {http://papers.nips.cc/paper\_files/paper/2023/hash/3d77c6dcc7f143aa2154e7f4d5e22d68-Abstract-Conference.html} {Learning to compress prompts with gist tokens}.
\newblock In \emph{Advances in Neural Information Processing Systems 36: Annual Conference on Neural Information Processing Systems 2023, NeurIPS 2023, New Orleans, LA, USA, December 10 - 16, 2023}.

\bibitem[{OpenAI(2023)}]{GPT4-report}
OpenAI. 2023.
\newblock \href {https://doi.org/10.48550/ARXIV.2303.08774} {{GPT-4} technical report}.
\newblock \emph{CoRR}, abs/2303.08774.

\bibitem[{Peng et~al.(2024)Peng, Quesnelle, Fan, and Shippole}]{YaRN}
Bowen Peng, Jeffrey Quesnelle, Honglu Fan, and Enrico Shippole. 2024.
\newblock \href {https://openreview.net/forum?id=wHBfxhZu1u} {Yarn: Efficient context window extension of large language models}.
\newblock In \emph{The Twelfth International Conference on Learning Representations, {ICLR} 2024, Vienna, Austria, May 7-11, 2024}. OpenReview.net.

\bibitem[{Qian et~al.(2024)Qian, Zhang, Liu, Mao, and Dou}]{memorag}
Hongjin Qian, Peitian Zhang, Zheng Liu, Kelong Mao, and Zhicheng Dou. 2024.
\newblock \href {https://doi.org/10.48550/ARXIV.2409.05591} {Memorag: Moving towards next-gen {RAG} via memory-inspired knowledge discovery}.
\newblock \emph{CoRR}, abs/2409.05591.

\bibitem[{Qin and Durme(2023)}]{Nugget}
Guanghui Qin and Benjamin~Van Durme. 2023.
\newblock \href {https://proceedings.mlr.press/v202/qin23a.html} {Nugget: Neural agglomerative embeddings of text}.
\newblock In \emph{International Conference on Machine Learning, {ICML} 2023, 23-29 July 2023, Honolulu, Hawaii, {USA}}, volume 202 of \emph{Proceedings of Machine Learning Research}, pages 28337--28350. {PMLR}.

\bibitem[{Qwen-Team(2024)}]{Qwen2}
Qwen-Team. 2024.
\newblock \href {https://doi.org/10.48550/ARXIV.2407.10671} {Qwen2 technical report}.
\newblock \emph{CoRR}, abs/2407.10671.

\bibitem[{Rae et~al.(2020)Rae, Potapenko, Jayakumar, Hillier, and Lillicrap}]{PG19}
Jack~W. Rae, Anna Potapenko, Siddhant~M. Jayakumar, Chloe Hillier, and Timothy~P. Lillicrap. 2020.
\newblock \href {https://openreview.net/forum?id=SylKikSYDH} {Compressive transformers for long-range sequence modelling}.
\newblock In \emph{8th International Conference on Learning Representations, {ICLR} 2020, Addis Ababa, Ethiopia, April 26-30, 2020}. OpenReview.net.

\bibitem[{Shazeer(2019)}]{MQA}
Noam Shazeer. 2019.
\newblock \href {https://arxiv.org/abs/1911.02150} {Fast transformer decoding: One write-head is all you need}.
\newblock \emph{CoRR}, abs/1911.02150.

\bibitem[{Sun et~al.(2024)Sun, Dong, Zhu, Huang, Wang, Ma, Zhang, Wang, and Wei}]{YOCO}
Yutao Sun, Li~Dong, Yi~Zhu, Shaohan Huang, Wenhui Wang, Shuming Ma, Quanlu Zhang, Jianyong Wang, and Furu Wei. 2024.
\newblock \href {https://doi.org/10.48550/ARXIV.2405.05254} {You only cache once: Decoder-decoder architectures for language models}.
\newblock \emph{CoRR}, abs/2405.05254.

\bibitem[{Suzgun et~al.(2023)Suzgun, Scales, Sch{\"{a}}rli, Gehrmann, Tay, Chung, Chowdhery, Le, Chi, Zhou, and Wei}]{BBH}
Mirac Suzgun, Nathan Scales, Nathanael Sch{\"{a}}rli, Sebastian Gehrmann, Yi~Tay, Hyung~Won Chung, Aakanksha Chowdhery, Quoc~V. Le, Ed~H. Chi, Denny Zhou, and Jason Wei. 2023.
\newblock \href {https://doi.org/10.18653/V1/2023.FINDINGS-ACL.824} {Challenging big-bench tasks and whether chain-of-thought can solve them}.
\newblock In \emph{Findings of the Association for Computational Linguistics: {ACL} 2023, Toronto, Canada, July 9-14, 2023}, pages 13003--13051. Association for Computational Linguistics.

\bibitem[{Tay et~al.(2020)Tay, Dehghani, Bahri, and Metzler}]{EfficientTransformer-Survey}
Yi~Tay, Mostafa Dehghani, Dara Bahri, and Donald Metzler. 2020.
\newblock \href {https://arxiv.org/abs/2009.06732} {Efficient transformers: {A} survey}.
\newblock \emph{CoRR}, abs/2009.06732.

\bibitem[{Wang et~al.(2024)Wang, Ma, Zhang, Ni, Chandra, Guo, Ren, Arulraj, He, Jiang, Li, Ku, Wang, Zhuang, Fan, Yue, and Chen}]{MMLU-Pro}
Yubo Wang, Xueguang Ma, Ge~Zhang, Yuansheng Ni, Abhranil Chandra, Shiguang Guo, Weiming Ren, Aaran Arulraj, Xuan He, Ziyan Jiang, Tianle Li, Max Ku, Kai Wang, Alex Zhuang, Rongqi Fan, Xiang Yue, and Wenhu Chen. 2024.
\newblock \href {https://doi.org/10.48550/ARXIV.2406.01574} {Mmlu-pro: {A} more robust and challenging multi-task language understanding benchmark}.
\newblock \emph{CoRR}, abs/2406.01574.

\bibitem[{Wei et~al.(2022)Wei, Wang, Schuurmans, Bosma, Ichter, Xia, Chi, Le, and Zhou}]{CoT}
Jason Wei, Xuezhi Wang, Dale Schuurmans, Maarten Bosma, Brian Ichter, Fei Xia, Ed~H. Chi, Quoc~V. Le, and Denny Zhou. 2022.
\newblock \href {http://papers.nips.cc/paper\_files/paper/2022/hash/9d5609613524ecf4f15af0f7b31abca4-Abstract-Conference.html} {Chain-of-thought prompting elicits reasoning in large language models}.
\newblock In \emph{Advances in Neural Information Processing Systems 35: Annual Conference on Neural Information Processing Systems 2022, NeurIPS 2022, New Orleans, LA, USA, November 28 - December 9, 2022}.

\bibitem[{Wu and Tu(2024)}]{LayerCondensedKVCache}
Haoyi Wu and Kewei Tu. 2024.
\newblock \href {https://doi.org/10.18653/V1/2024.ACL-LONG.602} {Layer-condensed {KV} cache for efficient inference of large language models}.
\newblock In \emph{Proceedings of the 62nd Annual Meeting of the Association for Computational Linguistics (Volume 1: Long Papers), {ACL} 2024, Bangkok, Thailand, August 11-16, 2024}, pages 11175--11188. Association for Computational Linguistics.

\bibitem[{Xiao et~al.(2024{\natexlab{a}})Xiao, Zhang, Han, Xiao, Lin, Zhang, Liu, Han, and Sun}]{InfLLM}
Chaojun Xiao, Pengle Zhang, Xu~Han, Guangxuan Xiao, Yankai Lin, Zhengyan Zhang, Zhiyuan Liu, Song Han, and Maosong Sun. 2024{\natexlab{a}}.
\newblock \href {https://doi.org/10.48550/ARXIV.2402.04617} {Infllm: Unveiling the intrinsic capacity of llms for understanding extremely long sequences with training-free memory}.
\newblock \emph{CoRR}, abs/2402.04617.

\bibitem[{Xiao et~al.(2024{\natexlab{b}})Xiao, Tang, Zuo, Guo, Yang, Tang, Fu, and Han}]{DuoAttention}
Guangxuan Xiao, Jiaming Tang, Jingwei Zuo, Junxian Guo, Shang Yang, Haotian Tang, Yao Fu, and Song Han. 2024{\natexlab{b}}.
\newblock \href {https://doi.org/10.48550/ARXIV.2410.10819} {Duoattention: Efficient long-context {LLM} inference with retrieval and streaming heads}.
\newblock \emph{CoRR}, abs/2410.10819.

\bibitem[{Yen et~al.(2024)Yen, Gao, Hou, Ding, Fleischer, Izsak, Wasserblat, and Chen}]{Helmet}
Howard Yen, Tianyu Gao, Minmin Hou, Ke~Ding, Daniel Fleischer, Peter Izsak, Moshe Wasserblat, and Danqi Chen. 2024.
\newblock \href {https://doi.org/10.48550/ARXIV.2410.02694} {{HELMET:} how to evaluate long-context language models effectively and thoroughly}.
\newblock \emph{CoRR}, abs/2410.02694.

\bibitem[{Zaheer et~al.(2020)Zaheer, Guruganesh, Dubey, Ainslie, Alberti, Onta{\~{n}}{\'{o}}n, Pham, Ravula, Wang, Yang, and Ahmed}]{bigbird}
Manzil Zaheer, Guru Guruganesh, Kumar~Avinava Dubey, Joshua Ainslie, Chris Alberti, Santiago Onta{\~{n}}{\'{o}}n, Philip Pham, Anirudh Ravula, Qifan Wang, Li~Yang, and Amr Ahmed. 2020.
\newblock \href {https://proceedings.neurips.cc/paper/2020/hash/c8512d142a2d849725f31a9a7a361ab9-Abstract.html} {Big bird: Transformers for longer sequences}.
\newblock In \emph{Advances in Neural Information Processing Systems 33: Annual Conference on Neural Information Processing Systems 2020, NeurIPS 2020, December 6-12, 2020, virtual}.

\bibitem[{Zellers et~al.(2019)Zellers, Holtzman, Bisk, Farhadi, and Choi}]{HellaSwag}
Rowan Zellers, Ari Holtzman, Yonatan Bisk, Ali Farhadi, and Yejin Choi. 2019.
\newblock \href {https://doi.org/10.18653/V1/P19-1472} {Hellaswag: Can a machine really finish your sentence?}
\newblock In \emph{Proceedings of the 57th Conference of the Association for Computational Linguistics, {ACL} 2019, Florence, Italy, July 28- August 2, 2019, Volume 1: Long Papers}, pages 4791--4800. Association for Computational Linguistics.

\bibitem[{Zhang et~al.(2024{\natexlab{a}})Zhang, Liu, Xiao, Shao, Ye, and Dou}]{Beacon}
Peitian Zhang, Zheng Liu, Shitao Xiao, Ninglu Shao, Qiwei Ye, and Zhicheng Dou. 2024{\natexlab{a}}.
\newblock Long context compression with activation beacon.
\newblock \emph{arXiv preprint arXiv:2401.03462}.

\bibitem[{Zhang et~al.(2024{\natexlab{b}})Zhang, Chen, Hu, Xu, Chen, Hao, Han, Thai, Wang, Liu, and Sun}]{Infbench}
Xinrong Zhang, Yingfa Chen, Shengding Hu, Zihang Xu, Junhao Chen, Moo~Khai Hao, Xu~Han, Zhen~Leng Thai, Shuo Wang, Zhiyuan Liu, and Maosong Sun. 2024{\natexlab{b}}.
\newblock \href {https://doi.org/10.48550/ARXIV.2402.13718} {{\(\infty\)}bench: Extending long context evaluation beyond 100k tokens}.
\newblock \emph{CoRR}, abs/2402.13718.

\bibitem[{Zhang et~al.(2024{\natexlab{c}})Zhang, Bo, Ma, Li, Chen, Dai, Zhu, Dong, and Wen}]{long-term-survey}
Zeyu Zhang, Xiaohe Bo, Chen Ma, Rui Li, Xu~Chen, Quanyu Dai, Jieming Zhu, Zhenhua Dong, and Ji{-}Rong Wen. 2024{\natexlab{c}}.
\newblock \href {https://doi.org/10.48550/ARXIV.2404.13501} {A survey on the memory mechanism of large language model based agents}.
\newblock \emph{CoRR}, abs/2404.13501.

\bibitem[{Zhang et~al.(2023)Zhang, Sheng, Zhou, Chen, Zheng, Cai, Song, Tian, R{\'{e}}, Barrett, Wang, and Chen}]{H2O}
Zhenyu Zhang, Ying Sheng, Tianyi Zhou, Tianlong Chen, Lianmin Zheng, Ruisi Cai, Zhao Song, Yuandong Tian, Christopher R{\'{e}}, Clark~W. Barrett, Zhangyang Wang, and Beidi Chen. 2023.
\newblock \href {http://papers.nips.cc/paper\_files/paper/2023/hash/6ceefa7b15572587b78ecfcebb2827f8-Abstract-Conference.html} {{H2O:} heavy-hitter oracle for efficient generative inference of large language models}.
\newblock In \emph{Advances in Neural Information Processing Systems 36: Annual Conference on Neural Information Processing Systems 2023, NeurIPS 2023, New Orleans, LA, USA, December 10 - 16, 2023}.

\bibitem[{Zhangir~Azerbayev()}]{Proof-Pile}
Bartosz~Piotrowski Zhangir~Azerbayev, Edward~Ayers.
\newblock \href {https://huggingface.co/datasets/hoskinson-center/proof-pile} {Proofpile: A pre-training dataset of mathematical texts.}

\bibitem[{Zhao et~al.(2023)Zhao, Zhou, Li, Tang, Wang, Hou, Min, Zhang, Zhang, Dong, Du, Yang, Chen, Chen, Jiang, Ren, Li, Tang, Liu, Liu, Nie, and Wen}]{llm-survey}
Wayne~Xin Zhao, Kun Zhou, Junyi Li, Tianyi Tang, Xiaolei Wang, Yupeng Hou, Yingqian Min, Beichen Zhang, Junjie Zhang, Zican Dong, Yifan Du, Chen Yang, Yushuo Chen, Zhipeng Chen, Jinhao Jiang, Ruiyang Ren, Yifan Li, Xinyu Tang, Zikang Liu, Peiyu Liu, Jian{-}Yun Nie, and Ji{-}Rong Wen. 2023.
\newblock \href {https://doi.org/10.48550/ARXIV.2303.18223} {A survey of large language models}.
\newblock \emph{CoRR}, abs/2303.18223.

\bibitem[{Zhou et~al.(2022)Zhou, Dou, Yuan, and Ma}]{socialformer}
Yujia Zhou, Zhicheng Dou, Huaying Yuan, and Zhengyi Ma. 2022.
\newblock \href {https://doi.org/10.1145/3485447.3511962} {Socialformer: Social network inspired long document modeling for document ranking}.
\newblock In \emph{{WWW} '22: The {ACM} Web Conference 2022, Virtual Event, Lyon, France, April 25 - 29, 2022}, pages 339--347. {ACM}.

\bibitem[{Zhu et~al.(2023)Zhu, Yuan, Wang, Liu, Liu, Deng, Dou, and Wen}]{irllm-survey}
Yutao Zhu, Huaying Yuan, Shuting Wang, Jiongnan Liu, Wenhan Liu, Chenlong Deng, Zhicheng Dou, and Ji{-}Rong Wen. 2023.
\newblock \href {https://doi.org/10.48550/ARXIV.2308.07107} {Large language models for information retrieval: {A} survey}.
\newblock \emph{CoRR}, abs/2308.07107.

\end{thebibliography}

\clearpage
\appendix
\section{Training Details}
\label{appendix: training details}
We train all models using 2B tokens from the upsampled SlimPajama dataset, with document boundaries marked by the eos token. Each model was augmented with 4 sink tokens to enhance modeling stability. To support dynamic compression ratio assignment, the compression ratio for each data instance is randomly sampled from \{4, 8, 16, 32\}. The context length of the training data is set to 16K, with a fixed segment length of 2K. The learning rate is set to 1e-5, using a cosine lr scheduler that reduces the learning rate to 50\% of its highest value in the end. Additionally, the first 1\% of training steps are allocated for learning rate warmup.

\section{Evaluation Details}
\label{appendix: evaluation details}
\paragraph{Perplexity}
The average perplexity is calculated across all data using a 16K-length context window, with a sliding window stride equal to the length of the context window.

\paragraph{Weak Context-dependent Tasks}
To ensure that the context for each task is compressed at least once, few-shot examples are used to fill the context. The number of examples used for each task is detailed in Table~\ref{table: evaluation setting of weak context-dependent tasks}. For all tasks except HellaSwag, which selects answers based on the likelihood of candidate answers, the Chain-of-Thought (CoT) reasoning approach is employed to generate answers.

\begin{table}[h]
    \centering
    \small
    \scalebox{1.0}{\begin{tabular}{c|cc}
        \toprule
        Dataset & \#Few-shot demos & Answer acquisition \\ 
        \midrule
        MMLU-Pro & 12 & Chain-of-Thought \\
        BBH & 8 & Chain-of-Thought \\
        GSM8K & 16 & Chain-of-Thought \\
        HellaSwag & 32 & Logits \\
        \bottomrule
    \end{tabular}}
        \caption{Evaluation setting of weak context-dependent tasks.}
        \label{table: evaluation setting of weak context-dependent tasks}
\end{table}

\begin{table}
    \centering
    \small
    \scalebox{0.9}{\begin{tabular}{ccc}
        \toprule
        Category & Tasks & Metrics \\
        \midrule
        \multirow{4}{*}{RAG} & NQ & SubEM \\
        & TriviaQA & SubEM \\
        & PopQA & SubEM \\
        & HotpotQA & SumEM \\
        \midrule
        Rerank & MS Marco & NDCG@10 \\
        \midrule
        \multirow{2}{*}{Long-doc QA} & $\infty$Bench QA & ROUGE Recall \\
        & $\infty$Bench MC & Accuracy \\
        \midrule
        \multirow{5}{*}{Many-shot ICL} & TREC Coarse & Accuracy \\
        & TREC Fine & Accuracy \\
        & NLU & Accuracy \\
        & BANKING77 & Accuracy \\
        & CLINIC150 & Accuracy \\
        \midrule
        \multirow{4}{*}{Synthetic recall} & JSON KV & SubEM \\
        & RULER MK Needle & SubEM \\
        & RULER MK UUID & SubEM \\
        & RULER MV & SubEM \\
        \midrule
        \multirow{2}{*}{Summ.} & $\infty$Bench Sum & ROUGE-Sum F1 \\
        & Multi-LexSum & ROUGE-Sum F1 \\
        \midrule
        Code & RepoBench & Edit Distance \\
        \bottomrule
    \end{tabular}}  
        \caption{Details of long context tasks.}
        \label{table: long context details}
\end{table}

\begin{table}
    \centering
    \small
    \scalebox{0.9}{\begin{tabular}{c|cccc}
        \toprule
        Type & MMLU-Pro & BBH & GSM8K & HellaSwag \\
        \midrule 
        Full Attention & 35.1 & 59.0 & 50.9 & 79.8 \\
        Coarse, Rec & 34.8 & 59.2 & 50.4 & 79.3 \\
        Coarse, KV & 35.1 & 58.5 & 51.6 & 79.2 \\
        Fine, KV & 35.0 & 59.5 & 50.1 & 79.5 \\
        \midrule
    \end{tabular}}
        \caption{Performance of short context tasks.}
        \label{table: short context tasks}
\end{table}

\begin{table*}
    \centering
    \small
    \begin{tabular}{c|c|ccccccc|c}
        \toprule
         Ratio & Compression Type & RAG & Rerank & LongQA & ICL & Synthetic & Summ. & Code & Average\\
        \midrule
        - & Full Attention& 56.2 & 26.6 & 44.5 & 67.1 & 81.8 & 19.0 & 64.6 & 51.4 \\
        \midrule
        \multirow{3}{*}{4} & Coarse-grained, Recurrent & 44.1 & 0.9 & 35.6 & 27.9 & 12.1 & 19.3 & 56.9 & 28.1 \\
        & Coarse-grained, KV Cache & 45.4 & 1.6 & 36.2 & 29.8 & 12.4 & 17.8 & 59.4 & 29.2 \\
        & Fine-grained, KV Cache & 54.8 & 10.6 & 43.8 & 67.5 & 15.5 & 18.2 & 59.4 & \textbf{38.9} \\
        \midrule
        \multirow{3}{*}{8} & Coarse-grained, Recurrent & 49.8 & 1.3 & 36.0 & 25.9 & 11.2 & 17.7 & 58.6 & 28.6 \\
        & Coarse-grained, KV Cache & 44.8 & 0.5 & 39.3 & 28.5 & 12.3 & 18.1 & 59.4 & 28.9 \\
        & Fine-grained, KV Cache & 52.0 & 5.0 & 44.2 & 62.7 & 11.6 & 17.9 & 61.7 & \textbf{36.4} \\
        \midrule
        \multirow{3}{*}{16} & Coarse-grained, Recurrent & 49.9 & 1.4 & 34.9 & 20.8 & 11.2 & 17.8 & 57.5 & 27.6 \\
        & Coarse-grained, KV Cache & 45.1 & 0.9 & 38.6 & 27.9 & 12.2 & 17.8 & 58.7 & 28.7 \\
        & Fine-grained, KV Cache & 49.5 & 3.1 & 42.2 & 44.5 & 11.7 & 16.9 & 59.6 & \textbf{32.5} \\
        \midrule
        \multirow{3}{*}{32} & Coarse-grained, Recurrent & 44.2 & 2.4 & 34.1 & 27.5 & 11.5 & 18.5 & 57.3 & 27.9 \\
        & Coarse-grained, KV Cache & 45.0 & 1.1 & 37.1 & 23.6 & 12.2 & 17.6 & 57.9 & 27.8 \\
        & Fine-grained, KV Cache & 47.5 & 1.7 & 40.6 & 36.9 & 12.1 & 16.8 & 59.5 & \textbf{30.8} \\
        \bottomrule
    \end{tabular}
    \caption{Long context performance based on \textsc{Qwen2-7B}.}
    \label{table: qwen2 long context result}
\end{table*}

\paragraph{Long Context Tasks} 
The majority of our task configurations are based on \citet{Helmet} and \citet{Prolong}, with code tasks leveraging RepoBench. We sample up to 1K samples for each dataset, and contexts are constructed under the configs of a max length of 16K. Details are presented in Table~\ref{table: long context details}. We apply greedy decoding to all generation tasks for stability.

\subsection{Results in the Short Context Setting}
\label{appendix: short context performance}
We report model performance in the short context setting in Table~\ref{table: short context tasks}, in which 2-shot demos are applied and contexts are not compressed. The results indicate that short-context capabilities are not affected by learning compression.

\section{Performance of Qwen2-7B}
\label{appendix: qwen performance}
In addition to \textsc{Llama3.1-8B}, we also conduct a full set of experiments on another widely acknowledged model, \textsc{Qwen2-7B}. The results are shown in Table~\ref{table: qwen2 long context result}.

\section{Results of Supervised Fine-tuning}
\label{appendix: sft results}
Supervised Fine-tuning (SFT) is a critical factor influencing model performance on downstream tasks. Gist token-based context compression models often struggle with certain tasks (e.g., synthetic ones), which may be attributed to the low proportion of long-dependency data in the general-purpose continue-training corpus. To investigate the effect of high-quality SFT data on the model’s compression ability, we fine-tune the \textsc{Llama3.1-8B-Instruct} with the Fine-KV architecture. The training data is consisted with LongAlpaca~\cite{LongLoRA}, BookSum~\cite{BookSum}, and synthetic data from~\cite{Beacon}. We then evaluate its performance on long-context tasks. Table~\ref{table: SFT results} presents the detailed results: the fine-tuned model shows significant gains in the previously weakest task (i.e., synthetic recall), while maintaining its performance on tasks where it already excelled. This suggests that long-range supervised signals effectively enhance the ability of gist tokens to preserve precise information in dense memory. Thus, high-quality SFT data containing long-distance dependencies is not only beneficial but potentially essential for the compression model.

\begin{table}
    \centering
    \small
    \scalebox{0.88}{\begin{tabular}{c|cccc|c}
        \toprule
         Compression Type & RAG & ICL & Synthetic & Summ. & Avg.\\ 
         \midrule
        Fine-KV& 59.9 & \textbf{75.5} & 54.1 & \textbf{21.0} & 52.6 \\
        + SFT & \textbf{60.2} & 73.3 & \textbf{66.3} & \textbf{21.7} & \textbf{55.4} \\
        \bottomrule
    \end{tabular}}
        \caption{Performance of the compression model after SFT (compression ratio=4).}
        \label{table: SFT results}
\end{table}

\section{Extrapolation Capabilities}
This work explores a segment-wise context compression method that can effectively reduce the maximum length that each transformer block needs to model. For example, taking \textsc{Llama3-8B} as an example, assuming a fixed compression ratio of 4 and a segment length of 1K, the context length after continue-training would be the same as the pre-training length, which is 8K. Even if the user’s input context length reaches 16K, exceeding the maximum length after continue-training, the actual maximum length that each transformer block needs to model would only be (16K-1K)/4+1K=4.75K, which still falls within the pre-trained context length of the model. Since the model has already learned the corresponding positional encodings during pre-training, this method holds promise for extrapolating actual inference lengths.

Using \textsc{Llama3.1-8B} as the base model, we evaluate the compressed model trained with 16K contexts on tasks involving 32K contexts. As shown in Table~\ref{table: extrapolation results}, the results indicate that the compressed model continues to perform well even with context lengths multiple times longer than the training length. This suggests that the ability to read context from gist tokens is generalizable.

\begin{table}
    \centering
    \small
    \scalebox{0.90}{\begin{tabular}{c|cc|ccc|c}
        \toprule
         Length & Model & CR. & RAG & ICL & Synthetic & Avg.\\ 
         \midrule
         \multirow{2}{*}{16K} & Full & - & 61.8 & 62.3 & 93.9 & 72.7 \\
         & Fine-KV & 4 & 60.4 & 72.7 & 62.1 & 65.1 \\
         \midrule
         \multirow{2}{*}{32K} & Full & - & 60.5 & 74.9 & 88.7 & 74.7 \\
         & Fine-KV & 4 & 59.3 & 76.8 & 34.9 & 57.9 \\
        \bottomrule
    \end{tabular}}
        \caption{Performance of compression models when inference length exceeds training length.}
        \label{table: extrapolation results}
\end{table}

\begin{table*}
\centering
\scalebox{0.9}{\begin{tabular}{p{3cm} p{12cm}}
\toprule
\multicolumn{2}{c}{\textbf{A Synthetic Example in PopQA}} \\
\midrule
\multicolumn{2}{c}{\textbf{Subject is relevant, and needle type is food}} \\
Subject: & John Peter Jukes \\
Document 1: & For the cartoonist with the same name see John Jukes. The Right Reverend John Peter Jukes (7 August 1923) was an English prelate of the Roman Catholic Church. He was a member of the Conventual Franciscans. Jukes was born in Eltham...\\
Document 2: & Richard Jukes was born on 9 October 1804 at Goathill, and died 10 August 1869. He served as a Primitive Methodist minister from 1827 to 1859. Jukes married Phoebe Pardoe in 1825, and later, widowed, he married Charlotte... \\
$\text{\color{blue}Golden doc:}$ & [Some content] $\text{\color{red}John Peter Jukes's special food is beef burger.}$ [The rest of content...] \\
More documents: & ... \\
Question: & What's the special food of John Peter Jukes? \\

\midrule
\multicolumn{2}{c}{\textbf{Subject is relevant, and needle type is number}} \\
Subject: & John Peter Jukes \\
Document 1: & For the cartoonist with the same name see John Jukes. The Right Reverend John Peter Jukes (7 August 1923) was an English prelate of the Roman Catholic Church. He was a member of the Conventual Franciscans. Jukes was born in Eltham...\\
Document 2: & Richard Jukes was born on 9 October 1804 at Goathill, and died 10 August 1869. He served as a Primitive Methodist minister from 1827 to 1859. Jukes married Phoebe Pardoe in 1825, and later, widowed, he married Charlotte... \\
$\text{\color{blue}Golden doc:}$ & [Some content] $\text{\color{red}John Peter Jukes's special number is 51681396.}$ [The rest of content...] \\
More documents: & ... \\
Question: & What's the special number of John Peter Jukes? \\

\midrule
\multicolumn{2}{c}{\textbf{Subject is irrelevant, and needle type is food}} \\
Subject: & John Peter Jukes \\
Document 1: & For the cartoonist with the same name see John Jukes. The Right Reverend John Peter Jukes (7 August 1923) was an English prelate of the Roman Catholic Church. He was a member of the Conventual Franciscans. Jukes was born in Eltham...\\
Document 2: & Richard Jukes was born on 9 October 1804 at Goathill, and died 10 August 1869. He served as a Primitive Methodist minister from 1827 to 1859. Jukes married Phoebe Pardoe in 1825, and later, widowed, he married Charlotte... \\
$\text{\color{blue}Golden doc:}$ & [Some content] $\text{\color{red}Mr. Tree's special food is beef burger.}$ [The rest of content...] \\
More documents: & ... \\
Question: & What's the special food of Mr. Tree? \\

\midrule
\multicolumn{2}{c}{\textbf{Subject is irrelevant, and needle type is number}} \\
Subject: & John Peter Jukes \\
Document 1: & For the cartoonist with the same name see John Jukes. The Right Reverend John Peter Jukes (7 August 1923) was an English prelate of the Roman Catholic Church. He was a member of the Conventual Franciscans. Jukes was born in Eltham...\\
Document 2: & Richard Jukes was born on 9 October 1804 at Goathill, and died 10 August 1869. He served as a Primitive Methodist minister from 1827 to 1859. Jukes married Phoebe Pardoe in 1825, and later, widowed, he married Charlotte... \\
$\text{\color{blue}Golden doc:}$ & [Some content] $\text{\color{red}Mr. Tree's special number is 51681396.}$ [The rest of content...] \\
More documents: & ... \\
Question: & What's the special number of Mr. Tree? \\

\bottomrule
\end{tabular}}
\caption{A synthetic example in PopQA for evaluate ``Lost if surprise''. The $\text{\color{red}Red}$ parts denote synthetic needles inserted to the dataset.}
\label{table: synthetic popqa example}
\end{table*}

\end{document}